\documentclass[10pt]{article} % For LaTeX2e
\usepackage[accepted]{tmlr}
\usepackage{amsmath,amsfonts,bm}

\def\eqref#1{equation~\ref{#1}}
\def\Eqref#1{Equation~\ref{#1}}
\def\1{\bm{1}}

\DeclareMathAlphabet{\mathsfit}{\encodingdefault}{\sfdefault}{m}{sl}
\SetMathAlphabet{\mathsfit}{bold}{\encodingdefault}{\sfdefault}{bx}{n}

\usepackage{hyperref}
\usepackage{url}
\usepackage{wrapfig}% standard LaTeX package
\allowdisplaybreaks

\usepackage{algorithm}
\usepackage{algorithmic}
\usepackage{amsmath}
\usepackage{amssymb}
\usepackage{booktabs}
\usepackage{graphicx}      % already loaded by most classes
\usepackage{subcaption}    % gives the subfigure environment
\usepackage{bbm}
\usepackage[final, commandnameprefix=none,authormarkup=none]{changes}
\definechangesauthor[name={Reviewer 6zg7 round}, color=orange]{R}
\definechangesauthor[name={Camera-ready}, color=red]{CR}

\title{ACTIVA: Amortized Causal Effect Estimation via Variational Autoencoders}

\author{\name Andreas Sauter \email a.sauter@vu.nl \\
      \addr Learning and Reasoning Group\\
      Vrije Universiteit Amsterdam
      \AND
      \name Saber Salehkaleybar \email s.salehkaleybar@liacs.leidenuniv.nl \\
      \addr Leiden Institute of Advanced Computer Science (LIACS)\\
      Leiden University
      \AND
      \name Frank van Harmelen \email frank.van.harmelen@vu.nl \\
      \addr Learning and Reasoning Group\\
      Vrije Universiteit Amsterdam
      \AND
      \name Aske Plaat \email a.plaat@liacs.leidenuniv.nl \\
      \addr Leiden Institute of Advanced Computer Science (LIACS)\\
      Leiden University
      \AND
      \name Erman Acar \email erman.acar@uva.nl \\
      \addr Informatics Institute \& Institute for Logic, Language and Computation\\
      University of Amsterdam}

\def\month{10}  % Insert correct month for camera-ready version
\def\year{2026} % Insert correct year for camera-ready version
\def\openreview{\url{https://openreview.net/forum?id=819pIIPuVl}} % Insert correct link to OpenReview for camera-ready version

\begin{document}

\maketitle

\begin{abstract}
Predicting post-intervention distributions from observational data is central to many scientific and decision-making problems, but remains challenging due to causal ambiguity, restrictive modeling assumptions, and the lack of amortization across tasks. We introduce ACTIVA, a transformer-based conditional variational autoencoder for amortized estimation of full interventional distributions from observational data and intervention queries. ACTIVA learns a conditional latent prior that amortizes causal knowledge across diverse training tasks, so that a new causal model drawn from the training distribution is handled in a single forward pass, without interventional data, a causal graph, or per-task fitting. Beyond the training distribution, only a parametric shift of the simulator is tested.
We provide a population-level characterization of ACTIVA's estimand, showing that, under idealized expressivity and infinite samples assumptions, ACTIVA's learning objective targets a simulator-weighted mixture over the interventional distributions of causal models that are observationally compatible with the input. Empirically, on synthetic datasets and biologically realistic gene-expression simulations, ACTIVA substantially outperforms a correlational baseline, reduces spurious non-descendant effects, and achieves competitive performance relative to strong amortized baselines. On a pair of observationally equivalent models sharing a single observational dataset, ACTIVA recovers the predicted mixture, validating the theory in a finite-capacity model. Our results show that ACTIVA is a promising approach for estimating interventional distributions from observational data.
\end{abstract}

\begin{figure}[htbp]
    
    \centering
    \includegraphics[width=\textwidth]{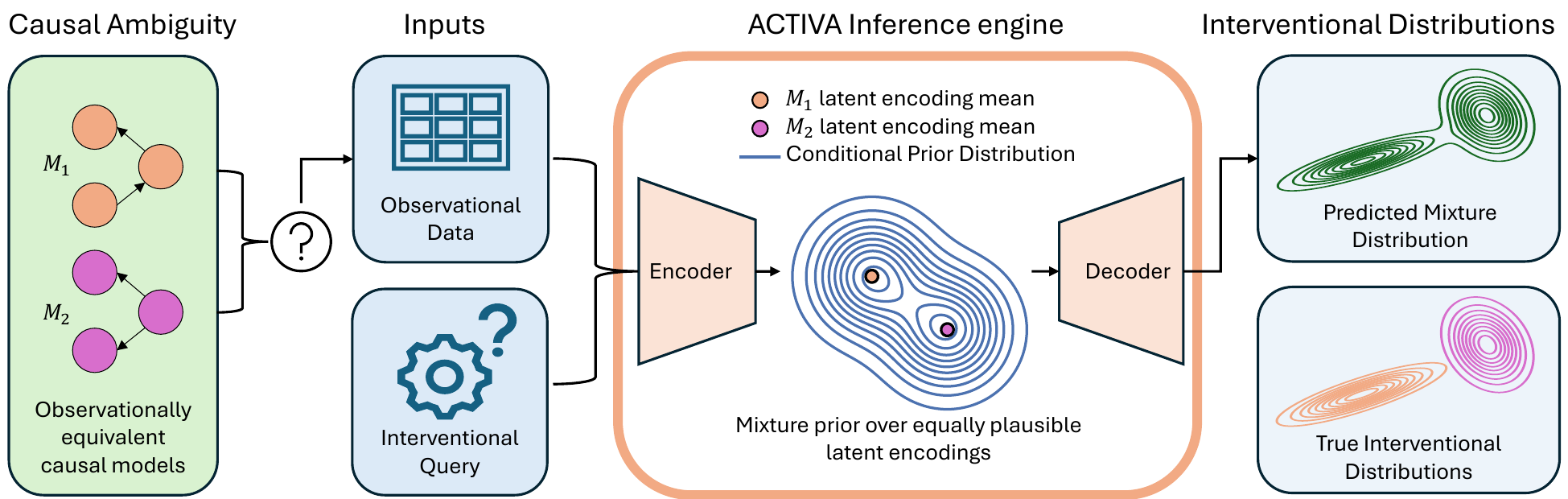}
    \caption{Causal models and their corresponding interventional distributions are generally ambiguous from observational data alone. Given observational data and an intervention query, ACTIVA encodes this ambiguity via a prior that represents the mixture over ambiguous models in the latent space. Through ACTIVA's decoder, the latent prior distribution is then transformed into a mixture over the interventional distributions of the ambiguous causal models. Section~\ref{sec:non_identifiability} demonstrates this behavior empirically on a pair of observationally equivalent models.%$\boldsymbol{D}^\mathcal{M}$ and a query intervention $do(X=x)$, the encoder maps to a latent point and the prior to a mixture of latent points. Samples $\boldsymbol{z}$ from this latent space are passed to the decoder, which estimates the interventional distribution $p_\mathcal{M}(\boldsymbol{V}\mid do(X=x))$ as a Gaussian mixture. Dashed lines indicate training-time information flow, solid lines represent inference-time.
    }
   \label{fig:graphical_abstract}
\end{figure}

\section{Introduction}
% What problem are we solving and why is it important
% Understanding the causal effects of an intervention based on observational data is central in many domains such as healthcare~\citep{Shi2022LearningSummary}, economics~\citep{Panizza2014PublicEffect}, and finance~\citep{Kumar2023CausalSurvey}, among others. However, estimating these effects is constrained by theoretical limits regarding the conclusions that can be drawn from observations only~\citep{Bareinboim2022OnInference}. Specifically, looking only at observational data leads to ambiguities about which causal effect a certain intervention will have.

Estimating what would happen under an intervention from observational data alone is a fundamental goal in fields such as healthcare, economics, and finance~\citep{Shi2022LearningSummary,Panizza2014PublicEffect,Kumar2023CausalSurvey}. These predictions are essential for decision-making, since in many real-world settings interventions are costly, risky, or impossible to test directly. Yet observational data by itself is often insufficient to determine causal effects uniquely: different causal mechanisms can generate the same observed data while implying different outcomes under intervention~\citep{Bareinboim2022OnInference}. This fundamental ambiguity makes causal effect estimation from observations alone both important and challenging.

 To address these ambiguities, current approaches often introduce additional assumptions, for instance by assuming access to the causal graph~\citep{Sanchez-Martin2022VACA:Queries,Sanchez2022DiffusionEstimation,Chao2023InterventionalModels,Poinsot2024LearningChallenges}. % or by restricting the functional form of the underlying causal model~\citep{Yang2021CausalVAE:Models,Qi2023CMVAE:Meta-Learning}. 
  An alternative approach to address these ambiguities is to make them explicit in the form of uncertainties. For example, providing an interval or distribution of plausible causal effects has been shown to be feasible and useful in practice ~\citep{Maathuis2009ESTIMATINGDATA, Balazadeh2025CausalPFN:Learning}. Such a prediction allows for a clearer picture of what range of effects to expect. % and can lead to actionable inferences if e.g. the range of effects lies within an interval that leads to the same actions. 
 Although having a distribution over causal effects is a promising direction, it only represents uncertainty over a specific point-estimate. 
 A natural extension is to look at the overall distributional shift under intervention of the causally relevant variables directly. % after performing an action. 
In addition to answering cause-effect point estimates, such a distribution can provide further task-relevant insights, such as expressing ambiguity via multi-modal distributions and answer questions of risk  while still providing point estimates about specific effects. For example, a treatment is judged not by its average effect but by the chance that a patient ends up in an acceptable state, with the symptom marker low enough to help and the side-effect marker not high enough to harm; that is the probability of a region under the joint, which no collection of per-variable averages can determine. % and hence prevent misleading average effect predictions.%~\citep{Rissanen2021AModels}. 
 
 A further practical complication is that, in many real-world applications, rather than knowing the fully specified intervention, one often observes only an intervention label together with the causal variable(s) that the intervention targets. For example, we may want to estimate the effect of administering a blood-pressure medication on patient outcomes without precise information about, e.g., the exact dosage or timing.  %A practically relevant example arises in genetic perturbation experiments. In such settings, experiments are often specified only through a perturbation label and a target gene rather than as a precise value-setting operation on observed expression~\citep{Qi2013RepurposingExpression}. Even conceptually similar perturbations can differ in strength and mechanism~\citep{Tuladhar2019CRISPR-Cas9-basedMisregulation}. Recent machine-learning approaches address this regime by modeling perturbation descriptions via latent interventions~\citep{Schneider2025GenerativeModeling} but are have to be learned again for each problem instance.
 In such a setting, many current causal inference approaches are not directly applicable, as they assume fully specified interventions.

 Adding to that, causal inference methods often require extensive computation for each new problem instance. Recent work shows that across a diverse set of causal inference tasks, these computations can be amortized~\citep{Lowe2022AmortizedData,Lorch2022AmortizedLearning,Scetbon2024AModeling,Sauter2024CORE:Learning,Mahajan2024Zero-ShotModels,Annadani2025AmortizedLearning,Robertson2025DO-PFN:ESTIMATION,Balazadeh2025CausalPFN:Learning,Ma2025FoundationNetworks,Dhir2025EstimatingMeta-Learning}. That means that a model is trained upfront --- with significant computational effort --- to solve the task, such that inference is computationally cheap by reusing learned knowledge. These results suggest a way towards foundation models for causal inference.  %Results from these approaches suggest that we can effectively amortize over datasets coming from different causal models for various downstream tasks. %In a similar setup, amortized causal effect estimation has been studied but with restrictions to scenarios with known causal graphs~\citep{Mahajan2024Zero-ShotModels} or covariates~\citep{Robertson2025DO-PFN:ESTIMATION,Balazadeh2025CausalPFN:Learning,Ma2025FOUNDATIONPREPRINT}.

% Shortcomings of current approaches
%Current approaches rely on strong assumptions such as knowing the causal graph, and parametric assumptions on the underlying causal model. Furthermore, most approaches don't amortize during training and have to be re-trained for each problem instance. % mention especially VACA \citep{Sanchez-Martin2022VACA:Queries}, causalNF \citep{Javaloy2023CausalPractice}, Zero-Shot causal models \citep{Mahajan}

In this paper, we propose ACTIVA, a conditional variational autoencoder (CVAE) model for amortized causal inference from observational data and intervention labels. %During training, the encoder  maps  interventional and observational data to a latent space conditioned on an interventional query of interest. The decoder transforms the latent encoding into the data-generating interventional distribution. 
For inference, we condition our prior on the observational data and the intervention query of interest to obtain a latent distribution corresponding to plausible causal models. The decoder transforms the prior distribution into a mixture over interventional distributions that are plausible under the observational data. %We provide a theoretical analysis of our latent and decoded space showing that our model estimates mixture distributions over equally plausible causal models. 
In empirical evaluations, we validate that ACTIVA can successfully recover post-interventional distributions and point estimates at inference time, even on novel instances with implicit uncertainty. In contrast to approaches that first recover or assume a causal graph, ACTIVA directly amortizes the prediction of post-interventional distributions, using the decoder to map latent representations of causal uncertainty into full distributional predictions. %This allows for the potential zero-shot transfer from simulated scenarios to the real-world ones without knowing the  causal relations, avoiding common pitfalls when relying on such graphs~\citep{Poinsot2024LearningChallenges}.
% Experimental results  
In summary, our contributions are as follows:

\begin{itemize}
\item We provide a CVAE model (ACTIVA) designed for amortized post-interventional distribution estimation in regimes with partially specified interventions from observational data.  %Our model takes an observational dataset and a query intervention as input, predicts a latent representation of the dataset and outputs an estimate of the respective interventional distribution .

\item We provide a population-level characterization of ACTIVA's estimand, showing that, under idealized conditions, ACTIVA's learning objective targets a mixture over the interventional distributions of observationally equivalent causal models, and validate this result empirically on a pair of observationally equivalent models.

\item We implement ACTIVA and show empirically that the theory-guided architecture learns useful post-interventional predictors in finite-sample settings: it substantially improves over a correlational baseline, reduces spurious non-descendant effects, and remains competitive with strong amortized baselines.
\end{itemize}

Overall, our work highlights the significance of amortized causal inference as a tool to overcome traditional hurdles in distributional causal effect estimation. The code repository is available at \url{https://github.com/sa-and/ACTIVA}; our data and trained models at \url{https://osf.io/5vebr/}.

The remainder of the paper is organized as follows: We start by introducing the most important background concepts and notation (Section~\ref{sec:background}). We then define ACTIVA (Section~\ref{sec:ACTIVA}) and provide a theoretical characterization (Section~\ref{sec:theory}). We then go into detail on how we implemented ACTIVA in practice (Section~\ref{sec:architecture}) and our experimental setup (Section~\ref{sec:experimental_setup}). In Section~\ref{sec:experiments} we provide our empirical results. Finally, we discuss the related work (Section~\ref{sec:related_work}) and provide a conclusion to our paper (Section~\ref{sec:conclusion}).

\section{Background and Notation}\label{sec:background}
\textbf{Conditional $\beta$-VAEs.}
Variational Autoencoders (VAEs) are generative models that define the joint distribution 
$ p_\theta(\boldsymbol{x}, \boldsymbol{z}) = p_\gamma(\boldsymbol{x} | \boldsymbol{z}) p_\eta(\boldsymbol{z}) $,
where $\boldsymbol{z}$ are latent variables governing the data $\boldsymbol{x}$ and $\theta=\{\gamma,\eta\}$ are the parameters determining the data generation. The marginal data likelihood $ p_\theta(\boldsymbol{x}) $ is optimized using the evidence lower bound (ELBO)~\citep{Kingma2013Auto-EncodingBayes}:
\begin{equation}
\log p_\theta(\boldsymbol{x}) \geq \displaystyle \mathop{\mathbb{E}}_{q_\phi(\boldsymbol{z} | \boldsymbol{x})} \left[ \log p_\gamma(\boldsymbol{x} | \boldsymbol{z}) \right] - \operatorname{KL}(q_\phi(\boldsymbol{z} | \boldsymbol{x}) \| p_\eta(\boldsymbol{z})),
\end{equation}
where $\phi$ parametrizes the encoding distribution $q$. The ELBO has a reconstruction term that ensures the model accurately reconstructs the data from the latent representation, and a Kullback-Leibler divergence (KL) term that regularizes the encoding distribution $ q_\phi(\boldsymbol{z} \mid \boldsymbol{x}) $ to approximate some prior $ p_\eta(\boldsymbol{z}) $.

Conditional VAEs (CVAEs)~\citep{Sohn2015LearningModels} extend this framework to allow for conditional generation given some auxiliary information $\boldsymbol{c}$. Here, we adopt a condition-dependent prior variant in which the auxiliary information affects generation only through the latent variable, i.e., $\boldsymbol{x}\!\perp\!\!\!\perp\!\boldsymbol{c}\mid\boldsymbol{z}$, %as in identifiable VAE formulations with auxiliary variables~\citep{Khemakhem2020VariationalFramework}, 
yielding the generative process:
\begin{equation}\label{eq:cond_gen}
    p_\theta(\boldsymbol{x},\boldsymbol{z}| \boldsymbol{c})=p_\gamma(\boldsymbol{x}|\boldsymbol{z})p_\eta(\boldsymbol{z}|\boldsymbol{c}).
\end{equation} %The joint distribution in this case factorizes as:
%$$p(\boldsymbol{x}, \boldsymbol{z} \mid \boldsymbol{c}) = p(\boldsymbol{x} \mid \boldsymbol{z}, \boldsymbol{c}) p(\boldsymbol{z}),$$
%and the ELBO is similarly adapted for the conditional setting. 
In $\beta$-VAEs~\citep{Higgins2017Beta-VAE:Framework}, the KL term is scaled by a hyperparameter $\beta$, leading to the $\beta$-weighted variational objective:
\begin{equation}\label{eq:cond_elbo}
\displaystyle \mathop{\mathbb{E}}_{q_\phi(\boldsymbol{z} \mid \boldsymbol{x}, \boldsymbol{c})} \left[ \log p_\gamma(\boldsymbol{x} | \boldsymbol{z}) \right] - \beta \operatorname{KL}(q_\phi(\boldsymbol{z} | \boldsymbol{x}, \boldsymbol{c}) \| p_\eta(\boldsymbol{z}|\boldsymbol{c})).
\end{equation}
The choice of $\beta$ governs the trade-off between accurately reconstructing the data and disentangling the latent representations. While $\beta > 1$ emphasizes disentanglement, in our setting, we use $\beta < 1$ to prioritize accurate reconstruction. Note that for $\beta \neq 1$ this objective is no longer a lower bound on the conditional log-likelihood $\log p_\theta(\boldsymbol{x}\mid\boldsymbol{c})$; it coincides with the conditional ELBO only at $\beta = 1$.

%\subsection{Identifiability of CVAEs}
%A central aspect in VAEs is the ability to uniquely identify the generative model. This unique identification ensures meaningful latent distributions and priors. For a conditional generative model as defined in \Eqref{eq:cond_gen} this means that for two parameter sets $\theta , \theta^*$, if their marginal conditional distribution of the respective models coincide, they must have the same parameters, or more formally:
%\begin{equation}\label{eq:identifiability}
%    p_\theta(\boldsymbol{x}| \boldsymbol{c})=p_{\theta^*}(\boldsymbol{x}|\boldsymbol{c})\implies \theta = \theta^*
%\end{equation}

%This property allows us to conclude that if a learned conditional marginal coincides with the true distribution, then the learned generative model corresponds to the true one. 

%\citept{Khemakhem2020VariationalFramework} showed that for a broad class of latent variable models, identifiability as defined in \Eqref{eq:identifiability} can be achieved up to a simple transformation. Specifically their main theorem implies that given a conditional generative model as defined in \Eqref{eq:cond_gen} with a conditionally factorial prior on the latent variables and additional mild assumptions, the parameters $\theta$ of the generative process are identifiable up to a linear invertible transformation and pointwise nonlinearities on the latents. Furthermore, if the prior is determined by two or more sufficient statistics, then the CVAE can be identified up to a permutation and pointwise nonlinearity of the latents.

\textbf{Causal Models.}
In this paper, we employ the notion of structural causal models (SCM) for describing causal data generating processes. For a detailed definition, we refer the reader to~\cite{Bareinboim2022OnInference}. 

We treat a causal model $\mathcal{M}$ as a generative process over $d$ variables $\boldsymbol{V}=\{V_1, ..., V_d\}$, denoting an assignment of these variables as $\boldsymbol{v} =\{v_1, ..., v_d\}$. The model is causal in the sense that each variable is generated as a function of its direct causes and an exogenous noise term. Specifically, any variable $V_j \in \boldsymbol{V}$ is determined by its direct causes with $V_j\leftarrow f_j(Pa_{V_j}, U_j)$, where $f_j$ is an arbitrary function of the direct causes $Pa_{V_j}$ (representing parent variables of $V_j$) and a noise term $U_j$. %We call two models equivalent, writing $\mathcal{M} = \mathcal{M'}$,  if all their parameters are equal. 
The causal graph $G^\mathcal{M}$ of model $\mathcal{M}$ encodes all parent relations via directed edges between the corresponding variables. Each model $\mathcal{M}$ induces a joint distribution $p_\mathcal{M}(\boldsymbol{V})$, called the observational distribution, and we denote a dataset of $N$ i.i.d. samples from this distribution as  $\boldsymbol{D}^\mathcal{M} := \boldsymbol{v}^\mathcal{M}_{1:N}$, where $\boldsymbol{v}^\mathcal{M}_n \sim p_\mathcal{M}(\boldsymbol{V})$ for $1\leq n \leq N $. We refer to the class of models $[\mathcal{M}^o]:=\{ \mathcal M: p_{\mathcal M}(\boldsymbol V) = p_{\mathcal M^o}(\boldsymbol V)\}$  as observationally equivalent models defined by some representative model $\mathcal{M}^o$. %All models with the same observational distribution form a so-called Markov equivalence class (MEC).

In a causal model, performing a so-called intervention $do(V = v)$ manipulates $\mathcal{M}$ such that the target variable $V$ is forced to take on the value $v$, regardless of $V$'s causes. Such an intervention results in an intervened model that we denote as $\mathcal{M}_{do(V=v)}$ or $\mathcal{M}_{do(V)}$ when $v$ is clear from the context. %We denote the variables of $\mathcal{M}_{do(V=v)}$ as $\boldsymbol{V} \mid do(V=v)$.  
%In this work, $v$ is one possible intervention value that the model does not see during inference, but for which the training data contains samples. 
$\mathcal{M}_{do(V)}$ induces a joint distribution $p_\mathcal{M}(\boldsymbol{V}\mid do(V))$ that we call the interventional or post-interventional distribution. Similarly to the observational case, we denote a dataset of $N$ i.i.d. samples from this distribution as  $\boldsymbol{D}^{\mathcal{M}_{do(V)}} := \boldsymbol{v}^{\mathcal{M}_{do(V)}}_{1:N}$, where $\boldsymbol{v}^{\mathcal{M}_{do(V)}}_n \sim p_\mathcal{M}(\boldsymbol{V}\mid do(V))$.  Although  we use this definition of interventions for clarity, our approach does not rely on the full specification of the intervention presented. The definition above is atomic, fixing $V$ to a single value. We implement the more general family of stochastic interventions, in which the mechanism of $V$ is replaced by a draw from a fixed distribution over values, the atomic case being recovered when that distribution is a point mass. Everything that follows applies verbatim to this family, since it refers only to the post-interventional distribution $p_\mathcal{M}(\boldsymbol{V}\mid do(V))$ that the mechanism induces and never to the value itself. We therefore keep writing $do(V)$ throughout and introduce our instantiation of the family in Section~\ref{sec:interv_rep}.

The purpose of this work is to estimate the interventional distribution $p_\mathcal{M}(\boldsymbol{V}\mid do(V))$ from an observational dataset $\boldsymbol{D}^\mathcal{M}$.
%TODO: insert formally what identifiability is
In general, identifying this distribution is not possible without additional assumptions or interventional data, not least because of the ambiguities introduced by observational equivalence~\citep{Bareinboim2022OnInference}. This fundamental property of causality naturally extends to our work, which we address in the theoretical analysis section of this work. %Literature has shown that these kinds of models are identifialbe only for datasets that are indentifiable and that the model has been trained on \citep{Montagna2024DemystifyingTransformers}

%Lastly, we point out that, while SCMs are useful for describing our approach and the theoretical analysis, we do not rely on any specific formalization of causal models in this work.

%Such assumptions include knowing the (partial) causal graph or assuming additive noise models with non-Gaussian noise as the true data-generating process~\citep{Shimizu2006AKerminen}.

%This fact gives rise to an equivalence class of causal graphs that are all admissible candidates for the true data-generating distribution of $p_\mathcal{M}(\boldsymbol{V})$.

%However, the data generating SCM can often be narrowed down to an equivalence class of models based e.g. on interventional equivalence.
%This work aims at alleviating this challenge by using interventional data at training time and exploiting the learned, amortized information at inference time without needing interventional data or causal graphs. %work. 

%TODO: find a good place to talk about equivariance in causal model and why it is important
%ATE

\section{ACTIVA}\label{sec:ACTIVA}
In this section, we outline a $\beta$-CVAE specification for amortized post-interventional distribution estimation called ACTIVA. In our setup, a dataset of observational samples and a query intervention serve as a condition for the following generative process:
\begin{equation}\label{eq:activa_gen_process}
    p_\theta(\boldsymbol{V}, \boldsymbol{z} | \boldsymbol{D}^\mathcal{M}, do(V))
    = p_\gamma(\boldsymbol{V} |  \boldsymbol{z})  p_\eta(\boldsymbol{z} | \boldsymbol{D}^\mathcal{M}, do(V))
\end{equation}
In this process, the observational data $\boldsymbol{D}^\mathcal{M}$ together with an intervention query of interest $do(V)$ is mapped to a distribution over latent points $\boldsymbol{z}$ via some prior $ p_\eta$. For each of these latents, the decoder distribution $ p_\gamma$ then parametrizes a distribution over the data variables $\boldsymbol{V}$, leading to a full joint distribution $p_\theta$ over latents and data variables.

Intuitively, the latent prior captures uncertainty over the causal model: the same observational data may be compatible with multiple data-generating causal models, each implying potentially different interventional distributions. As a result, the predicted post-interventional distribution can be seen as an estimation of the true distribution capturing the uncertainty about the causal model.  In Section~\ref{sec:theory}, we formalize this intuition as a population-level characterization of ACTIVA's estimand: we characterize the target of ACTIVA's learning objective under idealized conditions.

We assume a training set $\mathcal{\mathbf{D}}^{tr}=[\boldsymbol{D}^{\mathcal{M}_{do(V)}}_j, \boldsymbol{D}^\mathcal{M}_j, do(V)_j]_{j=1}^J$ with $J$ post-interventional datasets $\boldsymbol{D}^{\mathcal{M}_{do(V)}}$, $J$ observational datasets $\boldsymbol{D}^\mathcal{M}$, and $J$ queries  $do(V)$. 
In our setting, a sampled training task
$\mathcal{\mathbf{D}}^{tr}_j \sim p_{tr}( \mathbf{D}^{tr})=\int p_{tr}\bigl( \boldsymbol{D}^{\mathcal{M}_{do(V)}}, \boldsymbol{D}^\mathcal{M}\mid \mathcal{M}, do(V)\bigr)\,p_{tr}(\mathcal{M}, do(V))\,d\mathcal{M}$
is an element from the training data distribution according to a pre-defined joint distribution $p_{tr}(\mathcal{M}, do(V))$ over causal models and intervention queries. Only the two datasets are generated conditionally here; the query coordinate of the task is the one drawn from the joint, so no marginalization over $do(V)$ takes place. This predefined distribution can be interpreted as a simulator over the causal models, and the interventions performed on them. Just as the support of its model marginal delimits the causal models the analysis speaks about, the support of its intervention marginal delimits the queries that are admissible at inference.

To find the parameters $\gamma, \eta, \phi$, we maximize the $\beta$-weighted variational objective of \Eqref{eq:cond_elbo} using the conditional posterior  $q_\phi(\boldsymbol{z} | \mathcal{\mathbf{D}}^{tr}_j)$. Here, $\boldsymbol{D}^{\mathcal{M}_{do(V)}}_j$ plays the role of the input to be reconstructed and $[\boldsymbol{D}^\mathcal{M}_j, do(V)_j]$ as the auxiliary information, mapping to $\boldsymbol{x}$ and $\boldsymbol{c}$ in \Eqref{eq:cond_elbo}, respectively. Our overall learning objective is:
\begin{align}\label{eq:activa_elbo}
    \mathcal{L}(\gamma, \eta, \phi; \mathcal{\mathbf{D}}^{tr}_j)= \displaystyle \mathop{\mathbb{E}}_{q_\phi(\boldsymbol{z} | \mathcal{\mathbf{D}}^{tr}_j)} \left[ \log p_\gamma(\boldsymbol{D}^{\mathcal{M}_{do(V)}}_j | \boldsymbol{z}) \right] - \beta \operatorname{KL}(q_\phi(\boldsymbol{z} |\mathcal{\mathbf{D}}^{tr}_j) \parallel p_\eta(\boldsymbol{z}| \boldsymbol{D}^\mathcal{M}_j, do(V)_j)).
\end{align}

Because the interventional samples are independent samples, the likelihood factorizes as
$
p_\gamma\ (
\boldsymbol{D}^{\mathcal{M}_{do(V)}}_j \mid \boldsymbol z)
=
p_\gamma\ (
\boldsymbol{v}^{\mathcal{M}_{do(V)}}_{1:N;j}\mid \boldsymbol z)
=
\Pi_{i=1}^N
p_\gamma\ (
\boldsymbol{v}^{\mathcal{M}_{do(V)}}_{i;j}\mid \boldsymbol z)$. Furthermore, we use the reparameterization trick~\citep{Kingma2013Auto-EncodingBayes} for gradient estimation, allowing backpropagation through the stochastic sampling process of $q_\phi$. 

%Furthermore, we emphasize that this formulation does not rely on a fully specified functional form of the intervention; instead, it is sufficient if each $do(V)_j$ that results in the samples $\boldsymbol{D}^{\mathcal{M}_{do(V)}}_j$ is specified by a label and the variables it targets.

Importantly, this setup does not require the intervention to be fully specified functionally; it suffices to know an intervention label together with the variables targeted by $do(V)_j$, allowing application to partially defined intervention settings. Specifically, a label is taken to determine an intervention \emph{mechanism}, and not a single value. Variation of the value under one label is therefore accommodated by construction rather than excluded. Furthermore, the characterization is indifferent to the definition of the mechanism, since it refers only to the post-interventional distribution  it induces. %Writing $p_\mathcal{M}(\boldsymbol{V}\mid do(V))$ with a label in the query slot does presume that the label picks out one such mechanism. That is a matter of how we have chosen to write the result rather than a limit of the argument: were a label to denote several mechanisms, letting the mixture in Proposition~2 range over model--mechanism pairs compatible with the input, rather than over models alone, accommodates them under the same proof.  What we claim, then, is the narrow statement that inference does not require the mechanism to be specified.
%A long as the target and an identifier of the intervention are known, allowing the application to partially defined intervention setting. 

%For our final objective, we aim to amortize prediction over the training distribution of the causal model to successfully apply our model to data sets that were not included in the training data. 

Our goal is to amortize causal inference by training ACTIVA over a distribution of tasks that resemble those expected during test time as well as possible. More explicitly, we expect the tasks during test time to come from similar families of causal models and have the same interventions as during training.
Accordingly, optimizing the amortized objective amounts to maximizing the expectation of this objective over the distribution of the training datasets.
%$\mathcal{M} \sim p(\boldsymbol{\mathcal{M}}$ and consequently datasets sampled from these models datasets $\boldsymbol{D}^\mathcal{M}$ sampled from a distribution over causal models with $p(\boldsymbol{\mathcal{M}}$for a distribution over datasets $p(\boldsymbol{D}^\mathcal{M})$ the learning objective or our model is
\begin{equation}\label{eq:amortized_objective}
    \max_{\gamma, \eta,\phi} \quad \mathop{\mathbb{E}}_{\mathcal{\mathbf{D}}^{tr}_j \sim p_{tr}( \mathbf{D}^{tr})} \mathcal{L}(\gamma, \eta, \phi; \mathcal{\mathbf{D}}^{tr}_j)
\end{equation}
This objective optimizes the model to learn inference procedures across datasets generated by model--query pairs drawn from $p_{tr}(\mathcal{M}, do(V))$,
%even if the specific dataset was not included in the training set. 
beyond specific datasets observed in the training. 
%It furthermore implies that the more general the training distribution of datasets is, the more the model will generalize to new datasets
The broader the training distribution of the datasets, the better the model can generalize to new datasets during test-time inference~\citep{Montagna2024DemystifyingTransformers}.

\section{Theoretical Analysis}\label{sec:theory}
In this section, we characterize the population-level estimand of ACTIVA's learning objective. Under idealized expressivity and infinite samples assumptions, we show that ACTIVA targets the post-interventional distribution obtained by averaging over causal models that are observationally compatible with the input dataset. The result is formulated at the level of the joint simulator distribution over causal models and intervention queries and therefore concerns unseen inference tasks drawn from the same task-generating process, with the support of the intervention marginal delimiting the queries admissible at inference. When all models in the observational equivalence class agree on the queried interventional distribution, this target reduces to that single post-interventional distribution. %Otherwise, ACTIVA targets a simulator-weighted mixture over the post-interventional distributions of observationally equivalent models. 
Thus, the analysis does not provide a finite-sample guarantee or claim identification of the true causal model from observational data alone; instead, it makes precise the distribution toward which the population objective directs optimization.

We emphasize that the result is stated at the level of the joint distribution over models, datasets and queries, and assumes no structural relationship between a test model and the models seen during training beyond both being draws from the same $p_{tr}$. What the simulator does determine is the object being targeted, since it supplies the mixture weights and task support. Conversely, the analysis offers no quantitative guidance on how to construct $p_{tr}$, although Section~\ref{sec:data} and Appendix~\ref{apx:datasets} specify the model classes, noise distributions and graphs we instantiate in our experiments.

%In this section, we provide a consistency result for ACTIVA. Under idealized population-level assumptions, we characterize the post-interventional distribution that the learning objective targets. The result does not claim that ACTIVA recovers the post-interventional distribution on finite data; rather, it identifies the estimand toward which the optimization is directed. Our result is at the simulator distribution $p_{tr}(\mathcal{M})$ level which applies to unseen inference tasks drawn from the same distribution. Importantly, our goal is not to show that the true causal model is identified from observational data alone, but that under idealized conditions, ACTIVA averages over the post-interventional distributions of models that are observationally compatible with the input dataset.

% Throughout, let $\boldsymbol{D}^{\mathcal{M}^o}$ denote an observational dataset generated by some representative model $\mathcal{M}^o$, and let $do(V)$ denote a fixed intervention query. All equivalence classes below are understood with respect to the support of $p_{tr}(\mathcal{M})$.

\textbf{Assumption 1.} In the infinite sample limit, the observational dataset $\boldsymbol{D}^{\mathcal{M}^o}$ identifies the class of observationally equivalent causal models $[\mathcal{M}^o]$.

This assumption only requires that infinitely many observational samples determine which models in the support of the simulator distribution remain observationally equivalent to the input. In particular, in the infinite-sample limit, the observational dataset $\boldsymbol{D}^{\mathcal{M}^o}$ acts as a sufficient statistic for the equivalence class $[\mathcal{M}^o]$, so any function conditioned on $\boldsymbol{D}^{\mathcal{M}^o}$, including the learned prior $p_\eta(z |\boldsymbol{D}^{\mathcal{M}^o}, do(V))$, depends on the data only through $[\mathcal{M}^o]$. This justifies denoting conditioning on the dataset and the equivalence class interchangeably throughout the analysis below.

\textbf{Assumption 2.} There exist globally optimal parameters $(\gamma^*,\eta^*,\phi^*)$ of \Eqref{eq:amortized_objective} such that for every example
$\mathcal{\mathbf{D}}^{tr}=[\boldsymbol{D}^{\mathcal{M}_{do(V)}}, \boldsymbol{D}^{\mathcal{M}}, do(V)]$
in the support of $p_{tr}(\mathbf{D}^{tr})$,
\begin{equation}\label{eq:theory_realizability_opt_integral}
    \int p_{\gamma^*}(\boldsymbol{V}\mid \boldsymbol{z})\,
    q_{\phi^*}(\boldsymbol{z}\mid \mathcal{\mathbf{D}}^{tr})\,
    d\boldsymbol{z}
    =
    p_{\mathcal{M}}(\boldsymbol{V}\mid do(V)).
\end{equation}

In other words, at the population optimum, the encoder-decoder pair recovers the correct post-interventional distribution for every task in the support of the simulator. This assumption is well-grounded in our training setting: the encoder observes samples drawn directly from the true interventional distribution, which is therefore measured directly from the training data without requiring graphical criteria or additional assumptions~\citep{Pearl2009Causality, Bareinboim2022OnInference}. What this assumption does require however, is the joint closure of two standard gaps in amortized variational inference~\citep{Cremer2018InferenceAutoencoders}: an approximation gap, requiring sufficient per-task expressiveness of the model, and an amortization gap, requiring a single shared parameterization to attain these per-task optima across all tasks. %This is the analogue, in our setting, of the standard consistency guarantee for conditional neural processes (Garnelo et al., 2018).
We stress that this is a demanding requirement. For standard architectures both gaps are found to be non-zero in practice~\citep{Cremer2018InferenceAutoencoders}. Assumption~2 asks for both to vanish simultaneously and exactly, and to do so for \emph{every} task in the support of the simulator. 

A third idealization is implicit in the statement. The encoder receives finite samples, which do not in general determine the distributions they are drawn from, so the equality can only hold in the infinite-sample limit stated at the start of this section. The assumption should accordingly be read as delimiting the idealization under which the estimand is characterized, not as a property we expect our implementation to satisfy; whether a finite-capacity model approximates the characterized target closely enough to be useful is an empirical question, which we take up in Section~\ref{sec:experiments}.
%; Section 5 discusses how our finite implementation approximates it.

We next define the population-level aggregated posterior distribution over all simulator tasks that share the same observational equivalence class and intervention query:
\begin{equation}\label{eq:agg_posterior_population_integral}
    \bar{q}_{\phi}(\boldsymbol{z}\mid [\mathcal{M}^o], do(V))
    :=
    \int
    q_{\phi}(\boldsymbol{z}\mid \mathcal{\mathbf{D}}^{tr})\,
    p_{tr}(\mathcal{\mathbf{D}}^{tr}\mid [\mathcal{M}^o], do(V))\,
    d\mathcal{\mathbf{D}}^{tr}.
\end{equation}
That is, $\bar{q}_{\phi}$ averages the encoder outputs over all tasks from the simulator consistent with the queried observational equivalence class $[\mathcal{M}^o]$ and that correspond to the same intervention query.

\textbf{Assumption 3.} The conditional prior family $p_\eta(\boldsymbol{z}\mid \boldsymbol{D}^{\mathcal{M}}, do(V))$ is expressive enough to represent the aggregated posterior defined in \Eqref{eq:agg_posterior_population_integral}.

Following the standard ELBO decomposition for amortized variational models~\citep{Hoffman2016ELBOBound}, the expected KL term in \Eqref{eq:amortized_objective}, restricted to tasks with observational class $[\mathcal{M}^o]$ and query $do(V)$, contributes to the optimization of the prior only through
\begin{equation}\label{eq:KL_decomposed_population_integral}
    \operatorname{KL}\!\left(
        \bar{q}_{\phi^*}(\boldsymbol{z}\mid [\mathcal{M}^o], do(V))
        \,\middle\|\,
        p_{\eta}(\boldsymbol{z}\mid \boldsymbol{D}^{\mathcal{M}^o}, do(V))
    \right),
\end{equation}
up to terms that do not depend on $\eta$. A derivation is given in Appendix~\ref{apx:kl_decomposition}.

\textbf{Proposition 1.} Under Assumptions 1--3, the optimal conditional prior for an input $\bigl(\boldsymbol{D}^{\mathcal{M}^o}, do(V)\bigr)$ satisfies
\begin{equation}\label{eq:prior_equals_agg_population_integral}
    p_{\eta^*}(\boldsymbol{z}\mid \boldsymbol{D}^{\mathcal{M}^o}, do(V))
    =
    \bar{q}_{\phi^*}(\boldsymbol{z}\mid [\mathcal{M}^o], do(V)).
\end{equation}

\emph{Proof.}
The dependence of the  objective on the prior parameters $\eta$ is given by a KL term in \Eqref{eq:KL_decomposed_population_integral}. This KL divergence is always non-negative and equals zero if and only if its two arguments coincide almost everywhere. Since Assumption 3 guarantees that the prior family can represent the aggregated posterior, the optimum is attained exactly at
$
p_{\eta^*}(\boldsymbol{z}\mid \boldsymbol{D}^{\mathcal{M}^o}, do(V))
=
\bar{q}_{\phi^*}(\boldsymbol{z}\mid [\mathcal{M}^o], do(V))
$.
$\Box$

\textbf{Remark (role of $\beta$).} The weight $\beta>0$ in \Eqref{eq:activa_elbo} multiplies the only term through which $\eta$ enters the objective, so it rescales that term without moving its minimizer, and Proposition~1 holds for every $\beta>0$. Assumption~2 is stated for the optimum of the $\beta$-weighted objective that is actually trained, so Proposition~2 and the estimand it characterizes do not depend on $\beta$ either. What $\beta$ does affect is the solution attaining this estimand: it sets the balance between reconstruction and prior matching, so the optimal encoder and prior, as well as the finite-capacity solution reached in practice, depend on $\beta$.  Since the KL term is the only one that trains the prior used at inference, a small $\beta$ weakens that training signal. To counter this, checkpoints are therefore selected on the held-out interventional NLL of the prior-driven predictive, which measures inference-time quality directly (Section~\ref{sec:training_and_inference}).

Proposition~1 shows that, at the population optimum, the learned prior does not select a single latent code. Instead, it matches the average encoder distribution over all simulator tasks that are observationally compatible with the input and share the same intervention query. An important structural feature of this formulation deserves emphasis: during training, the encoder $q_{\phi}(\boldsymbol{z}\mid \mathcal{\mathbf{D}}^{tr})$ observes both the observational and interventional samples, whereas at inference, only the prior $p_\eta(\boldsymbol{z}\mid \boldsymbol{D}^{\mathcal{M}}, do(V))$ receives the observational data and the intervention identifier alone. Thus, training concerns a strictly more informative setting than inference. The role of Proposition 1 is precisely to bridge this gap: it shows that the learned prior, optimized to match the aggregated encoder distribution, absorbs the information that the encoder extracted from interventional data during training and makes it available at inference time. We now characterize the resulting predictive distribution.

\textbf{Proposition 2.} Under Assumptions 1--3, the ACTIVA predictive distribution for an input $\bigl(\boldsymbol{D}^{\mathcal{M}^o}, do(V)\bigr)$ is given by
\begin{equation}\label{eq:main_mixture_result_integral}
    p_{\theta^*}(\boldsymbol{V}\mid \boldsymbol{D}^{\mathcal{M}^o}, do(V))
    =
    \int
    p_{\mathcal{M}}(\boldsymbol{V}\mid do(V))\,
    p_{tr}(\mathcal{M}\mid [\mathcal{M}^o], do(V))\,
    d\mathcal{M}.
\end{equation}

\emph{Proof sketch.}
Starting from the generative model in \Eqref{eq:activa_gen_process}, we first marginalize out the latent variable \(\boldsymbol z\). Substituting the result from Proposition~1 into the marginal shows that the predictive distribution is an average, with respect to \(p_{tr}(\mathcal{\mathbf D}^{tr}\mid [\mathcal M^o], do(V))\), over the task-wise decoder marginals induced by \(q_{\phi^*}(\boldsymbol z\mid \mathcal{\mathbf D}^{tr})\). Assumption~2 identifies each such decoder marginal with the true post-interventional distribution of the model that generated the corresponding task. Therefore, the overall prediction is exactly the simulator-weighted mixture of post-interventional distributions over models in the observational equivalence class \([\mathcal M^o]\). The full proof is given in Appendix~\ref{apx:prop2_proof}.

Proposition~2 characterizes the estimand of ACTIVA's learning objective: under idealized conditions, the objective targets a mixture over all post-interventional distributions that are observationally compatible with the input dataset and supported by the simulator distribution. The mixture weights are given explicitly by the conditional simulator distribution
$
p_{tr}(\mathcal{M}\mid [\mathcal{M}^o], do(V))
$,
which measures how much probability mass the training distribution assigns to each model among those that remain observationally compatible with the input and correspond to the queried intervention. Thus, the uncertainty represented by ACTIVA is not arbitrary but is induced jointly by observational ambiguity and by the joint simulator distribution $p_{tr}(\mathcal{M}, do(V))$.

\textbf{Corollary 1.} 
If all models in the observational equivalence class of the input agree on the queried interventional distribution, i.e., $p_{\mathcal{M}}(\boldsymbol{V}\mid do(V))=p_{\mathcal{M}^o}(\boldsymbol{V}\mid do(V))$ for all $\mathcal{M}\in[\mathcal{M}^o]$, then the queried effect is identifiable within the support of $p_{tr}(\mathcal{M}, do(V))$, and Proposition~2 reduces to
\begin{equation}
    p_{\theta^*}(\boldsymbol{V}\mid \boldsymbol{D}^{\mathcal{M}^o}, do(V))
    =
    p_{\mathcal{M}^o}(\boldsymbol{V}\mid do(V)).
\end{equation}

\emph{Proof.}
The conditional simulator distribution $p_{tr}(\mathcal{M}\mid [\mathcal{M}^o], do(V))$ is supported on $[\mathcal{M}^o]$, and by assumption the integrand $p_{\mathcal{M}}(\boldsymbol{V}\mid do(V))$ is constant and equal to $p_{\mathcal{M}^o}(\boldsymbol{V}\mid do(V))$ on that support.
Hence, the integral in \Eqref{eq:main_mixture_result_integral} collapses to
$
p_{\mathcal{M}^o}(\boldsymbol{V}\mid do(V))
$, since the conditional simulator distribution integrates to one.
$\Box$

Corollary~1 shows that whenever the queried effect is identifiable across the equivalence class, ACTIVA recovers the unique post-interventional distribution rather than a non-trivial mixture. Note that this does not require the equivalence class to be a singleton: collapse is query-specific, and a class containing several distinct models still yields a point prediction as long as they agree under the queried intervention. The singleton case $[\mathcal{M}^o]=\{\mathcal{M}^o\}$ is the special case in which the condition holds for every query.

Overall, this result provides a characterization of our training target and frames ACTIVA as an amortized estimator of post-interventional distributions under causal ambiguity. In identifiable settings, the objective targets the queried interventional distribution exactly; otherwise, it targets a principled mixture over the post-interventional distributions of observationally equivalent models supported by the simulator. Whether finite-sample, finite-capacity implementations approximate this target well is an empirical question, which we address in Section~\ref{sec:experiments}. Section~\ref{sec:non_identifiability} provides a direct check of both propositions: on a pair of observationally equivalent models, the learned prior is shown to match the aggregated posterior as in Proposition~1 and to decode into the simulator-weighted mixture as in Proposition~2.

\section{Model Architecture}\label{sec:architecture}
\begin{figure}
    \centering
    \includegraphics[width=\textwidth]{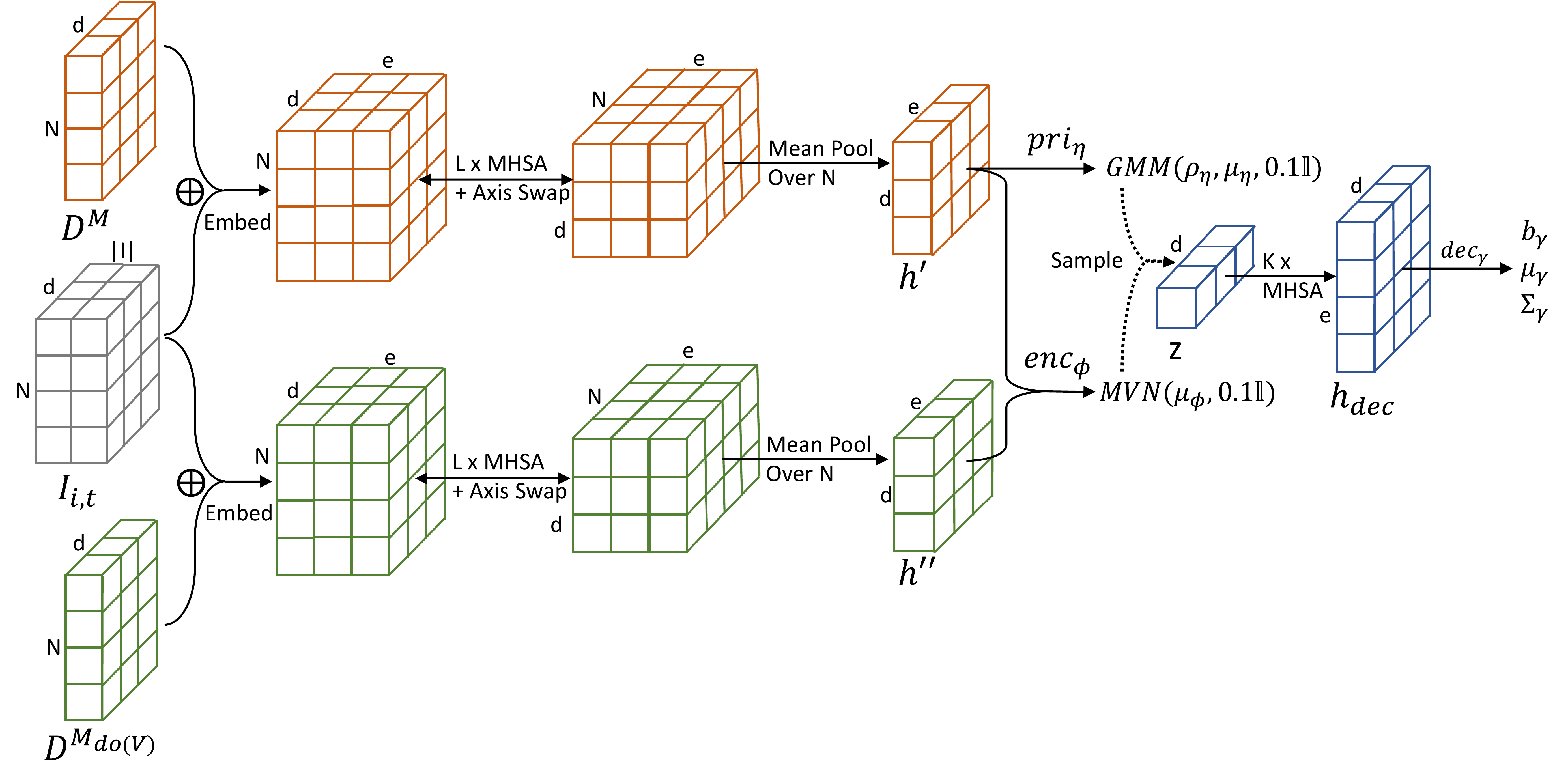}
    \caption{Overview of the proposed model architecture. %The model consists of three main processing pathways. 
    The orange blocks represent the observational stream, which processes observational data alongside the intervention representation to parameterize both the prior  and the posterior. The green blocks represent the interventional stream, which processes interventional samples exclusively during training to inform the posterior. Finally, the blue blocks represent the decoder stream. In a brief walkthrough: inputs are embedded, processed via alternating MHSA blocks, and mean-pooled over the sample dimension to yield intermediate representations ($h'$ and $h''$). These representations parameterize the GMM prior and MVN posterior. A latent variable $z$ is then sampled and passed through the transformer-based decoder to predict the final post-interventional mixture parameters ($b_\gamma$, $\mu_\gamma$, $\Sigma_\gamma$).}
    \label{fig:architecture}
\end{figure}

This section outlines the neural architecture of our proposed model, detailing how we jointly encode observational data and targeted interventions into a permutation-equivariant latent space. We subsequently describe the specific parameterizations of our prior and posterior distributions, as well as the transformer-based decoder used to generate post-interventional predictions. Figure~\ref{fig:architecture} provides an overview of the model architecture.

\textbf{Representing Interventions.}\label{sec:interv_rep}
To represent interventions, we create a matrix representation. We consider $i \in \{1, \dots, |I|\}$ the index of a possible intervention value $v_i$, where $|I|$ is the number of possible values, and a selector $\boldsymbol{t} \in \{0, 1\}^d$, with $d$ the number of variables, indicating the intervention target(s). By not encoding the intervention value directly, but rather the index of the intervention, we remove the need to have a full specification of the intervention. This allows us to model interventional distributions based on the index of the interventions, as long as they can be attributed to interventional training data and have specified targets. 

We construct the intervention representation $\boldsymbol{I}_{i,t}$ representing $do(\boldsymbol{V}_{\boldsymbol{t}} = v_i)$ as follows.
We perform a one-hot encoding of $i$ creating a vector $\boldsymbol{i_{oh}} \in \mathbb{R}^{\mid I \mid}$ and repeat it $d$ times, resulting in a matrix $\boldsymbol{i_{rep}} \in \mathbb{R}^{d \times \mid I \mid}$. We then apply $\boldsymbol{t}$ as a mask to this matrix, effectively zeroing out the rows that correspond to non-intervened variables. Finally, we repeat the intervention representation $N$ times to match the number of input samples and obtain $\boldsymbol{I}_{i,t} \in \mathbb{R}^{N \times d \times \mid I \mid}$. This construction ensures that the intervention-relevant information for each variable is provided as local information alongside the variable itself and maintains permutation equivariance.

Two properties of this construction deserve to be stated explicitly. First, $|I|$ is fixed when the network is built, so an intervention outside the set fixed at construction has no representation and the model cannot be queried with it. This is the architectural counterpart of a restriction already present in \Eqref{eq:amortized_objective}, where the support of the intervention marginal delimits the admissible queries. Second, per our model definition in Section~\ref{sec:ACTIVA}, an index $i$ need not stand for a single value, but for a mechanism. The mechanisms we instantiate are mostly stochastic rather than atomic  (Appendix~\ref{apx:linear_datasets}), so that a single label already covers a range of latent values. Nothing in the analysis depends on the shape of the distribution over values a mechanism induces.

The model receives no input beyond the samples and this intervention representation. In particular, there is no channel for structural knowledge such as a partial graph, a causal ordering, or forbidden edges, so such knowledge cannot be supplied at inference without retraining. This is a deliberate choice that reflects the most restrictive setting, in which no such information is available. In principle, such knowledge can be incorporated by retraining with it as an additional input to the condition, for instance as an adjacency matrix whose entries mark each edge as present, absent, or unknown, or a causal ordering. The simulator can generate this matrix from the graph of each training model, with entries masked as unknown so that the model learns to handle partial knowledge. Propositions~1 and~2 then carry over unchanged, with the mixture restricted to the models in $[\mathcal{M}^o]$ consistent with the supplied knowledge.

\noindent\textbf{Embedding Network.}
We encode the input data and intervention according to an embedding network $h_\alpha(\cdot)$ based on the extension of non-parametric encoders \citep{Kossen2021Self-AttentionLearning,Lorch2022AmortizedLearning}, where $\alpha$ are the parameters of the network. For various causal tasks, this architecture has been successfully used to encode datasets into a vector containing causally relevant information~\citep{Lorch2022AmortizedLearning,Scetbon2024AModeling,Annadani2025AmortizedLearning,Balazadeh2025CausalPFN:Learning,Robertson2025DO-PFN:ESTIMATION,Ma2025FoundationNetworks}. In particular, given some data   $\boldsymbol{D} \in \mathbb{R}^{N \times d}$ of $N$  samples, an intervention index $i$ and a binary target vector $\boldsymbol{t}$, we append $\boldsymbol{I}_{i,t}$ to the dataset, resulting in an augmented dataset $\boldsymbol{D}_{it} \in \mathbb{R}^{N \times d \times \mid I\mid + 1}$. Then we apply $L$ blocks of multi-head self-attention (MHSA) that alternate in attending over $d$ features and $N$ samples. After the transformer blocks, we average over the sample axis to obtain an embedding $\boldsymbol{h} \in \mathbb{R}^{d\times e}$, where $e$ indicates the embedding dimension, thus ensuring that the embedding is permutation invariant with respect to the sample dimension and equivariant with respect to the feature dimension.  

\textbf{Prior and Posterior.}\label{sec:prior_encoder}
Guided by our theoretical findings, we choose the conditional prior to be a  Gaussian Mixture Model (GMM) with $d$ dimensions and $C$ components. The component means $\boldsymbol{\mu}_\eta \in \mathbb{R}^{C \times d}$ and component weights $\boldsymbol{\rho}_\eta \in \mathbb{R}^C$ are determined by a learnable prior function $pri_\eta(h_{\alpha'}(\boldsymbol{D}^\mathcal{M}, \boldsymbol{I}_{i,t}))$ applied to the observational data. Specifically, $pri_\eta$ are two  multi-layer perceptrons (MLP) that we apply to the embedding $\boldsymbol{h}$ for each feature independently to compute the means and mixture weights. The outputs parameterize our prior distribution  $p_\eta(\boldsymbol{z}|\boldsymbol{D}^\mathcal{M}, do(V))  =\sum_{c=1}^{C}\rho_{\eta,c}\mathcal{MVN}(\boldsymbol{z};\boldsymbol{\mu_{\eta,c}}, 0.1\mathbbm{1}_d)$, where $\mathbbm{1}_d$ is the d-dimensional identity matrix. We set the prior to have a fixed variance of $0.1$ and no covariance for model simplicity, and enforce $\sum_{c=1}^{C}\rho_{\eta,c}=1$ via a softmax layer.

Furthermore, we choose the posterior distribution to be a $d$-dimensional multivariate normal distribution (MVN) with a constant, diagonal covariance matrix whose means are determined by a learnable encoder function $\boldsymbol{\mu}_\phi = enc_\phi(h_{\alpha''}(\boldsymbol{D}^{\mathcal{M}_{do(V)}}, \boldsymbol{I}_{i,t}), h_{\alpha'}(\boldsymbol{D}^\mathcal{M}, \boldsymbol{I}_{i,t}))$, where $h_{\alpha'}$ is the observational data network used in the prior, and $h_{\alpha''}$ is the network encoding interventional data with different parameters. We concatenate the resulting embeddings $\boldsymbol{h}'$ and $\boldsymbol{h}''$ along the embedding dimensions and again apply a two-layer MLP to each feature independently to compute the mean. This fully parametrizes our encoding distribution $q_\phi(\boldsymbol{z}|\boldsymbol{D}^{\mathcal{M}_{do(V)}},\boldsymbol{D}^\mathcal{M}, do(V)) = \mathcal{MVN}(\boldsymbol{z}; \boldsymbol{\mu}_\phi, 0.1\mathbbm{1}_d)$, with diagonal covariance matrix, matching the prior setup.

\noindent\textbf{Decoder. }\label{sec:decoder}
We model the decoder $dec_\gamma(\boldsymbol{z})$ as a transformer that outputs the parameters of a Gaussian mixture, enabling a closed-form expression for the estimated interventional distributions. We process $\boldsymbol{z} \in \mathbb{R}^d$ via $K$ standard transformer blocks~\citep{Vaswani2017AttentionNeed}, resulting in the embedding $\boldsymbol{h}_{dec} \in \mathbb{R}^{d \times e}$. We feed $\boldsymbol{h}_{dec}$ through a row-wise linear layer to predict the means  $\boldsymbol{\mu}_{\gamma} \in \mathbb{R}^{B\times d}$, where $B$ is the number of decoder mixture components, maintaining permutation equivariance regarding the variable ordering. To  predict the covariances $\boldsymbol{\Sigma}_{\gamma} \in \mathbb{R}^{B \times d \times d}$ in a permutation equivariant manner, we first apply a row-wise linear layer to compute $\boldsymbol{u} \in \mathbb{R}^{B \times d \times e}$ from $\boldsymbol{h}_{dec}$, and then compute the covariances via $\boldsymbol{\Sigma}_{\gamma} = \boldsymbol{u} \cdot \boldsymbol{u}$ similarly to \citep{Lorch2022AmortizedLearning}, and add a small constant $\epsilon$ to the diagonals for computational stability.
 Lastly, we estimate the decoder component weights $\boldsymbol{b}_{\gamma} \in \mathbb{R}^B$ of the mixture by first summing $\boldsymbol{h}_{dec}$ along the $d$ feature dimensions and then passing the pooled representation through a linear layer and a softmax layer.

\noindent \textbf{Permutation Equivariance.}
Importantly, we maintain permutation equivariance of the predicted mixture with respect to variable ordering. This is crucial because variable labels in causal problems are arbitrary: graphs that differ only by a relabeling represent the same underlying structure, and permutation equivariance forces the model to treat them consistently. As a result, the model can share statistical strength across all labelings of the same graph, reducing the effective graph hypothesis space by collapsing relabelings of the same underlying graph into a single case. Encoding this symmetry has been shown to improve statistical efficiency, scaling, and predictive performance in causal discovery~\citep{Lorch2022AmortizedLearning,Li2020SupervisedDiscovery}.

\noindent\textbf{Practical Approximation of the Theoretical Model.}
The theory in Section~\ref{sec:theory} guides the design of our finite-capacity implementation. In the population setting, ACTIVA predicts post-interventional distributions by averaging over simulator-supported causal models that are compatible with the observational input. We implement this idea by using a multi-modal conditional prior over latent representations and a decoder that maps latent samples to post-interventional mixture distributions. Since the idealized construction cannot be implemented exactly, we make several tractable approximations: finite observational samples replace population-level equivalence classes, a finite GMM prior approximates the induced latent mixture, and a Gaussian posterior with fixed diagonal covariance replaces the idealized  posterior. These approximations are chosen to preserve the central interpretation of the theory while enabling efficient training and inference. Empirically, the results in Section~\ref{sec:experiments} show that this theory-guided approximation is effective in practice: even when implemented through tractable finite approximations, ACTIVA learns predictive distributions that reflect post-interventional structure rather than merely observational correlations.

\section{Experimental Setup}\label{sec:experimental_setup}
This section outlines the experimental setup designed to assess our approach for predicting post-interventional distributions. We first detail the synthetic and semi-synthetic datasets, followed by a description of the baseline models, the specific evaluation protocol, and our implementation details for training and inference.

\subsection{Data}\label{sec:data}
We evaluate the performance of the proposed method on three types of datasets:  one biologically plausible type and two synthetic types. The synthetic datasets allow us to systematically analyze how well our amortized approach recovers linear causal effects under different noise assumptions. The semi-synthetic SERGIO datasets bring our method closer to real-world conditions by simulating biologically realistic gene-expression data under interventions. Further details on dataset generation are provided in  Appendix~\ref{apx:datasets}.

For the synthetic data, we generate samples from linear additive causal models with either Gaussian or Beta noise, in both 2-variable and 8-variable settings. We apply interventions on each variable and obtain four datasets in total: \emph{Gauss 2}, \emph{Beta 2}, \emph{Gauss 8}, and \emph{Beta 8}. We note that both settings are identifiable from observational data~\citep{Shimizu2006AKerminen,peters2014identifiability}.

For the semi-synthetic data, we use the SERGIO simulator to generate gene expression data that closely resembles real single-cell gene expression patterns~\citep{Dibaeinia2020SERGIO:Networks}. We simulate single-variable interventions corresponding to gene knockouts on each variable and construct datasets with 8 variables, as well as a controlled simulator-mismatch set, referred to as out-of-distribution (OOD) below, which has 11 variables instead of 8 and draws its simulator parameters from ranges disjoint or partially disjoint from the training ones. SERGIO's generative structure is held fixed throughout, so the shift is parametric within a fixed model class rather than a change of generative family.

\subsection{Baselines}\label{sec:baselines}
  %We compare our method against three baselines that use observational data to  approximate the interventional distribution.

 \noindent\textbf{Gaussian Conditional.}\;
  This baseline fits a multivariate Gaussian to the observational samples and uses the resulting  conditional distribution $ p(\boldsymbol{V}\setminus \boldsymbol{V}_{\boldsymbol{t}} \mid \boldsymbol{V}_{\boldsymbol{t}} = v_i) $ as a surrogate  for the interventional distribution $ p(\boldsymbol{V}\setminus \boldsymbol{V}_{\boldsymbol{t}} \mid do(\boldsymbol{V}_{\boldsymbol{t}} = v_i)) $. % For sample-based metrics and plotting that need the joint over all variables,  it assigns the intervention value $v_i$ to each variable in  $\boldsymbol{V}_{\boldsymbol{t}}$.  
  This baseline serves as a weak baseline on estimation accuracy that exploits all the correlation structure present  in the data but ignores the causal directionality.

  \noindent\textbf{Linear Causal.}\;
This baseline assumes access to the true causal graph and fits a linear SCM by ordinary least-squares regression of each variable on its parents~\citep{Bollen1989StructuralVariables}. For an intervention $\mathrm{do}(\boldsymbol{V}_{\boldsymbol{t}} = v_i)$, it computes $p(\boldsymbol{V}\setminus \boldsymbol{V}_{\boldsymbol{t}} \mid \mathrm{do}(\boldsymbol{V}_{\boldsymbol{t}} = v_i))$ analytically by graph surgery~\citep{Pearl2009Causality}, replacing the mechanism for $\boldsymbol{V}_{\boldsymbol{t}}$ with $v_i$ and propagating the resulting moments through the fitted linear system. Under correct specification, this baseline asymptotically recovers the exact interventional mean regardless of the Gaussianity of the noise. However, the full post-intervention distribution is Gaussian only under Gaussian noise; otherwise, the Gaussian prediction is misspecified, even if its mean is correct~\citep{Peters2017ElementsAlgorithms}. This baseline serves as an asymptotic oracle estimator of the post-interventional distributions for the Gaussian datasets, and a privileged-information estimator for the other datasets.

  \noindent\textbf{Nonlinear Causal.}\;
Like the Linear Causal baseline, this baseline assumes access to the true causal graph, but it replaces the linear assumption with a general additive noise model. We include it because the SERGIO mechanisms are neither linear nor Gaussian. For each evaluation task, a causal mechanism is assigned to every node by the automatic, cross-validated assignment procedure of \texttt{dowhy.gcm}~\citep{Blobaum2024DoWhyGCM}: root nodes receive an empirical distribution, and every non-root node an additive noise model $V_j = f_j(\mathrm{pa}(V_j)) + N_j$, with $f_j$ selected per node from a set of candidate regressors. The post-interventional distribution then follows from the g-formula for a known DAG~\citep{Robins1986ANewApproach,Pearl2009Causality}: the intervened node's factor is removed from the observational factorization and the remaining conditionals are carried over unchanged, so that $p(\boldsymbol{V}\setminus \boldsymbol{V}_{\boldsymbol{t}} \mid \mathrm{do}(\boldsymbol{V}_{\boldsymbol{t}} = v_i)) = \prod_{j \notin \boldsymbol{t}} p(V_j \mid \mathrm{pa}(V_j))\big|_{\boldsymbol{V}_{\boldsymbol{t}} = v_i}$, where each $p(V_j \mid \mathrm{pa}(V_j))$ is the observational conditional, any intervened parent is fixed to $v_i$, and each factor is evaluated as the fitted noise density at the residual. Like ACTIVA, it receives only the observational sample set of the task at hand, and it is fitted independently on every evaluation task. This factorization admits no closed-form marginal over an arbitrary subset of coordinates, so the descendant condition is instead evaluated with a density fitted to the baseline's own interventional samples and marked accordingly in Table~\ref{tab:main_results}. Since its mechanisms are estimated and do not match the generative form of the data, this baseline is a privileged-information estimator rather than an oracle.

  \noindent\textbf{MACE-TNP extension.} \;
  This model is our extension of MACE-TNP~\citep{Dhir2025EstimatingMeta-Learning}, a recent strong approach for post-interventional distribution prediction. We extend the original model formulation to predict the joint post-interventional distribution instead of variable-specific marginals, matching our problem specification. This  transformer-based neural process (TNP) uses an encoder to process observational context data and an interventional query, similar to our approach. The main high-level difference from our method is that it requires the exact specification of the intervention value during inference, constituting a slightly privileged information setting compared to ACTIVA, which only needs the identifier. The details of this model can be found in Appendix~\ref{app:mace-tnp-implementation}.                             

\subsection{Evaluation Protocol}\label{sec:marginal_subsets} 
We evaluate our model and baselines based on the following marginals. \textbf{All} includes all variables $\boldsymbol{V}$ including $\boldsymbol{V_t}$ (the intervened variable itself). This subset informs us about the overall predicted post-interventional distribution quality. \textbf{All$\setminus \boldsymbol{V_t}$} excludes the intervention variable. This allows for a separate evaluation for scenarios where the exact specification of the intervention is known, and hence the prediction of the intervened variable is not of interest. \textbf{Descendants} retains only variables that are descendants of $\boldsymbol{V_t}$ in the causal graph i.e. $\boldsymbol{V_t}$ is a direct or indirect cause, allowing for investigation of the prediction fit of true causal effects. $\boldsymbol{V_t}$ is not included, and descendants are identified via breadth-first search on the ground-truth adjacency matrix;  \textbf{Non-Descendants} retains all variables that are not descendants of $\boldsymbol{V_t}$ (excluding $\boldsymbol{V_t}$), that is, variables for which the true causal effect is zero by definition. Our evaluation metrics are as follows:

 \noindent\textbf{Average Treatment Effect (ATE).}\;
  We define the ATE of intervention $do(\boldsymbol{V}_{\boldsymbol{t}}=v_i)$ on $\boldsymbol{V}$ as
  \begin{equation}\label{eq:ate}
      \mathrm{ATE}%_{j \to y}
          \;:=\; \mathbb{E}\bigl[\boldsymbol{V} \mid do(\boldsymbol{V}_{\boldsymbol{t}}=v_i)\bigr]
              - \mathbb{E}\bigl[\boldsymbol{V} \mid do(\emptyset)\bigr],
  \end{equation}
  where the second expectation is taken under the observational distribution.
  We estimate the ground-truth ATE from held-out interventional samples and
  measure accuracy via mean absolute error (MAE) between the predicted and
  ground-truth ATEs.
  Because true ATEs are exactly zero for non-descendants, the
  non-descendant-MAE quantifies the rate of spurious predictions.

 \noindent\textbf{Negative Log-Likelihood (NLL).}\;
  We assess the calibration of the predicted interventional distribution by
  computing the average negative log-likelihood of the held-out interventional
  samples under the predicted density. %:
  % \begin{equation}
  %     \mathrm{NLL}
  %         \;=\; -\frac{1}{N}\sum_{n=1}^{N}
  %               \log\, p\bigl(\boldsymbol{v}^{\mathcal{M}_{do(\boldsymbol{V}_{\boldsymbol{t}})}}_n
  %                   \mid do(\boldsymbol{V}_{\boldsymbol{t}} = v_i)\bigr),
  % \end{equation}
  % where $\boldsymbol{v}^{\mathcal{M}_{do(\boldsymbol{V}_{\boldsymbol{t}})}}_n$ are held-out samples from the true
  % interventional distribution and $p$ is the density.
   NLL rewards both accurate means and well-calibrated variances.

 \noindent\textbf{Energy distance (ERG).}\;
  To assess distributional fidelity in the sample space, we
  compute the energy distance~\citep{Szekely2013EnergyDistances} between the set of
  predicted samples $\hat{\boldsymbol{v}}^{\mathcal{M}_{do(\boldsymbol{V}_{\boldsymbol{t}})}}_{1:N}$ and the held-out interventional samples
  $\boldsymbol{v}^{\mathcal{M}_{do(\boldsymbol{V}_{\boldsymbol{t}})}}_{1:N}$. %We define the energy distance as:
  % \begin{equation}
  %     \mathcal{E}(\hat{\boldsymbol{v}}^{\mathcal{M}_{do(\boldsymbol{V}_{\boldsymbol{t}})}}_n,\, \boldsymbol{v}^{\mathcal{M}_{do(\boldsymbol{V}_{\boldsymbol{t}})}}_n)
  %         \;=\; 2\,\mathbb{E}\bigl\|\hat{v}^{\mathcal{M}_{do(\boldsymbol{V}_{\boldsymbol{t}})}}_n - v^{\mathcal{M}_{do(\boldsymbol{V}_{\boldsymbol{t}})}}_n\bigr\|
  %             - \mathbb{E}\bigl\|\hat{v}^{\mathcal{M}_{do(\boldsymbol{V}_{\boldsymbol{t}})}}_n - {\hat{v}^{\mathcal{M}_{do(\boldsymbol{V}_{\boldsymbol{t}})}}_n}'\bigr\|
  %             - \mathbb{E}\bigl\|v^{\mathcal{M}_{do(\boldsymbol{V}_{\boldsymbol{t}})}}_n - {v^{\mathcal{M}_{do(\boldsymbol{V}_{\boldsymbol{t}})}}_n}'\bigr\|,
  % \end{equation}
  % where expectations are over independent draws from the respective sample sets.
  The energy distance is zero if and only if the two distributions are identical
  and is sensitive to differences in any moment of the distribution, providing a
  complementary view to NLL.

\subsection{Training and Inference}\label{sec:training_and_inference}
We train our model on the \emph{Beta}, \emph{Gaussian}, and \emph{SERGIO} datasets described above. For the 8 variable problems, we train our models for 1000 epochs; for the two variable problems and SERGIO we train 5000 epochs. We evaluate the NLL on the validation set every 25 and 100 steps, respectively, and retain the best performing model as our trained model. For all training runs, we used 20 mixture components for the prior GMM, 4 samples from the posterior to estimate the KL term during training, and 5 samples from the prior to create the mixture of the decoder during training. Architecturally, we used 4 blocks for the encoder and the decoder,  4 attention heads in each attention module, a hidden dimension of 256 for all modules, and an embedding dimension of 64. Regarding the other hyperparameters, we used a $\beta$ of 0.5 and a learning rate of .0005. All reported results use $\beta = 0.5$; the sensitivity of the results to $\beta$ has not been tested.  All modules and training scripts were implemented with Jax \citep{Bradbury2018JAX:Programs} and Flax \citep{Heek2024Flax:JAX}. Experiments were run on the DAS6 computing cluster \citep{bal16DAS}.

To infer the post-interventional distribution from our trained model, we follow the following procedure. First, the observational data and interventional query are provided to the prior as input, resulting in the prior distribution over latents. Since the prior is a GMM, obtaining closed-form solutions for interventional distributions at inference is intractable. We instead approximate the distribution by sampling 10 latent variables. We then decode each of these latent samples and form a uniform mixture of the resulting decoded distributions. Since we set the number of decoder components to 2 throughout our models, the resulting GMM has 20 components, matching our MACE-TNP extension. In the case of sample-based analysis, we sample from the predicted decoder distribution.

\section{Experiments}\label{sec:experiments}
We evaluate ACTIVA along four complementary axes. First, we compare predicted interventional samples of all models and baselines qualitatively to identify characteristic failure modes. Second, we quantify distributional fit using negative log-likelihood (NLL) and energy distance across different causal marginals. Third, we assess whether these distributional predictions translate into accurate downstream treatment-effect estimates via the ATE. Lastly, we validate Proposition 1 and 2 empirically.

\subsection{Qualitative Analysis}\label{sec:qualitative_results}
\begin{figure}[h]
    \centering
    \includegraphics[width=0.99\textwidth]{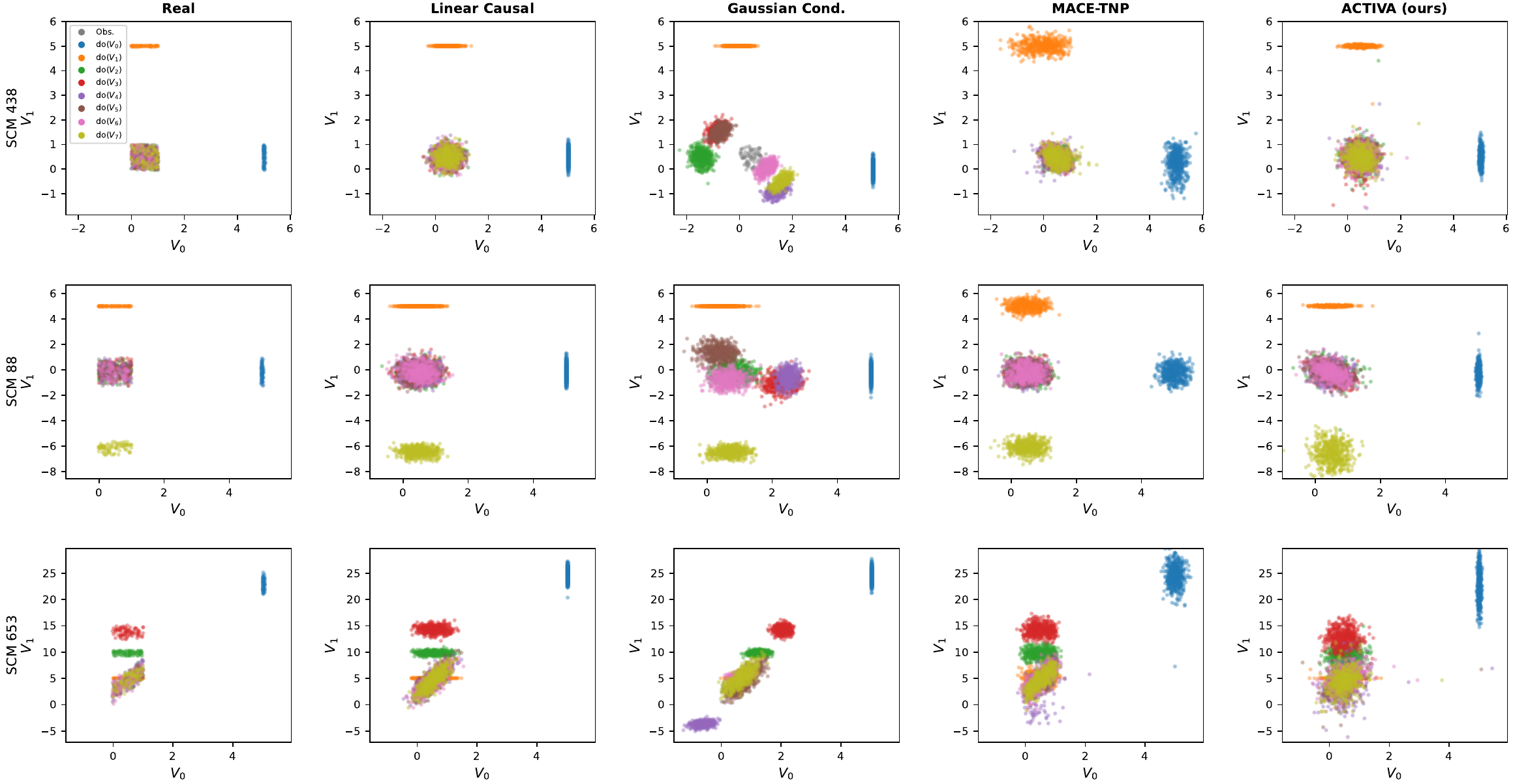}
    \caption{Samples from the observational and interventional distributions of the ground-truth model (first column), the baselines (columns 2-4), and our model (last column) for the \emph{Beta 8} data. Each row corresponds to a randomly sampled SCM from the test data. The colors indicate which variables have been intervened on ($do(V_i = 5)$). All samples are projected on the $V_0$ vs. $V_1$ plane. The closer the distributions are to those in the first column, the more accurate the prediction is.}
    \label{fig:8var_beta_comparison}
\end{figure}

We begin with a qualitative comparison of the predicted interventional distributions to better understand the behavior of each method. For four randomly selected test SCMs, we visualize samples from the predicted interventional distributions for all possible interventions. Figure~\ref{fig:8var_beta_comparison} shows the results for the Beta 8 dataset\footnote{The grey points are the ground-truth observational samples from the data, repeated in every plot for visualization.}; the corresponding plots for the remaining synthetic datasets are provided in Appendix~\ref{apx:distribution_plots}.

The Linear Causal baseline reproduces the main interventional shifts when its modeling assumptions hold, as expected, given access to the ground-truth graph. In contrast, the Gaussian Conditional baseline often predicts visible distributional changes even when no causal effect is present, illustrating the failure of a purely correlational predictor in this setting.

Comparing ACTIVA and MACE-TNP to the ground-truth samples, both methods recover the mean shifts induced by the interventions. Qualitatively, MACE-TNP sometimes matches the spread of the non-intervened variables more closely, whereas ACTIVA often yields tighter predictions for the intervened coordinates in the examples. Overall, these plots support the following quantitative results: ACTIVA captures the main interventional shifts while avoiding many of the spurious effects produced by the correlational baseline.

\subsection{Interventional Distribution Estimation}
\label{sec:main_results}
We next quantify how well ACTIVA captures the post-interventional distributions observed qualitatively above. For each test SCM, we run inference with the model trained on the corresponding training split and evaluate the resulting distributions using NLL and energy distance. For the Linear Causal and Gaussian Conditional baselines, we omit the \textit{All} condition because these methods do not directly predict the intervened dimensions. Bold and italic numbers indicate the best and second-best results per condition, respectively. Evaluation of MACE-TNP on the OOD SERGIO dataset is not possible because our implementation requires the same input dimensionality during training and inference.

\begin{table}[t]
    \centering
    \caption{Average inference performance of our trained model and the baseline on held-out test according to various marginals explained in Section~\ref{sec:marginal_subsets}. The \emph{Data} column indicates which dataset has been used with how many variables. We report the average energy distance and average negative log likelihood over the causal models in the corresponding test sets. The Linear Causal and Nonlinear Causal approaches have privileged information (PI) by having access to the causal graph; the Linear Causal approach is additionally an asymptotic oracle (O) for the Gaussian datasets. Entries marked $^{*}$ are obtained from a density fitted to the baseline's own interventional samples, since the g-formula factorization admits no closed-form marginal over the descendant subset. Lower is better for all metrics.}
    \label{tb:perf_comparison}
    
    % \scriptsize % No longer needed, table fits nicely!
    \begin{tabular}{@{}ll cccccccc@{}}
    \toprule
    & & \multicolumn{2}{c}{\textbf{All}}& \multicolumn{2}{c}{\textbf{All$\setminus \boldsymbol{V_t}$}} & \multicolumn{2}{c}{\textbf{Desc.}} & \multicolumn{2}{c}{\textbf{Non-Desc.}} \\
    \cmidrule(lr){3-4} \cmidrule(lr){5-6} \cmidrule(lr){7-8}\cmidrule(lr){9-10}
    \textbf{Data} & \textbf{Approach} & ERG $\downarrow$ & NLL $\downarrow$ & ERG $\downarrow$ & NLL $\downarrow$ & ERG $\downarrow$ & NLL $\downarrow$ & ERG $\downarrow$ & NLL $\downarrow$ \\
    \midrule
    
    Gauss 2
    & Linear Causal (O)&---&---&\textbf{0.09}&\textbf{0.90}&\textbf{0.32}&\textbf{1.39}&\textbf{0.02}&\textbf{0.73}\\
    & ACTIVA (ours)   &\textbf{0.14}&\textbf{-1.57}& \textit{0.14} & \textbf{0.90} & 0.48 & \textit{1.41} & \textit{0.03} & \textit{0.78} \\ % gallant eon
    & MACE-TNP &\textit{0.16}&-0.73& 0.17 & \textit{1.09} & \textit{0.47} & \textbf{1.39} & 0.06 & 0.99 \\
    & Gaussian Cond. &---&---& 1.16 & 8.55 & 0.33 & \textbf{1.39} & 1.43 & 11.00 \\
    \midrule
    
    Beta 2
    & Linear Causal (PI)&---&---&0.16&0.82&0.31&1.44&\textbf{0.01}&\textit{0.19}\\
    & ACTIVA   (ours) &\textbf{0.02}&\textbf{-2.27}&\textbf{0.02}&\textbf{0.22}&\textbf{0.03}&\textbf{0.24}&\textbf{0.01}&\textbf{0.18}\\ %lively spaceship
    & MACE-TNP &\textit{0.07}&\textit{-1.59}&\textit{0.07}&\textit{0.52}&\textit{0.13}&\textit{0.82}&\textbf{0.01}& 0.22\\
    & Gaussian Cond.&---&--- &2.00&37.8&0.31&1.43&3.69&74.14 \\
    \midrule
    
    Gauss 8
    & Linear Causal (O)&---&---&\textbf{0.34}&\textbf{5.82}&\textbf{0.53}&\textbf{3.46}&\textbf{0.08}&\textbf{4.08}\\
    & ACTIVA  (ours)  &\textit{1.84}&\textbf{6.65}&1.84&9.12&2.75&5.52&0.46&5.85 \\ %peach sun
    & MACE-TNP &\textbf{1.58}&\textit{8.79}&\textit{1.57}&\textit{8.57}&\textit{2.53}&\textit{5.18}&\textit{0.38}&\textit{5.48} \\
    & Gaussian Cond. &---&---&4.83&38.07&2.74&6.94&4.03&36.00 \\
    \midrule
    
    Beta 8
    & Linear Causal (PI)&---&---&\textbf{0.34}&\textbf{2.85}&\textbf{0.58}&\textbf{3.02}&\textbf{0.05}&\textbf{1.11}\\
    & ACTIVA   (ours) &\textit{1.43}&\textbf{3.06}&1.43&5.53&2.28&4.60&0.27&2.75 \\ %youthful resonance
    & MACE-TNP &\textbf{1.03}&\textit{4.25}&\textit{1.02}&\textit{4.47}&\textit{1.68}&\textit{3.77}&\textit{0.18}&\textit{2.11} \\
    & Gaussian Cond. &---&---&5.21&78.04&3.04&11.51&4.28&75.52 \\
    \midrule
    
    SERGIO
    & Linear Causal (PI)&---&---&\textbf{1.07}&14.07&\textit{1.88}&10.80&\textbf{0.78}&12.60\\
    & Nonlinear Causal (PI)&---&---&\textit{1.11}&14.83&2.05&12.15$^{*}$&\textit{0.79}&13.33\\
    & ACTIVA    (ours)&\textbf{1.14}&\textbf{6.70}&1.14&\textbf{9.23}&\textbf{1.86}&\textbf{3.94}&0.87&\textbf{8.99} \\ %comfy sea
    & MACE-TNP &\textit{1.20}&\textit{10.93}&1.17&\textit{10.36}&2.31&\textit{7.88}&0.80&\textit{9.31} \\
    & Gaussian Cond. &---&---& 2.81 & 28.85 & 1.97 & 15.12 & 2.67 & 26.92  \\
    \midrule
    
    OOD
    & Linear Causal (PI)&---&---&\textbf{1.07}&\textit{18.55}&\textbf{2.55}&\textit{16.96}&\textbf{0.73}&\textbf{15.96}\\
    & Nonlinear Causal (PI)&---&---&\textit{1.44}&20.31&4.65&28.95$^{*}$&\textit{0.74}&16.64\\
    & ACTIVA   (ours) &2.97&15.50&2.96&\textbf{17.97}&7.63&\textbf{11.43}&1.87&\textit{16.48} \\ %comfy sea
    & Gaussian Cond.&---&---& 8.22 & 90.87 &\textit{3.00}&23.29&8.20&88.56   \\
    \bottomrule
    \end{tabular}
    \label{tab:main_results}
\end{table}

% \noindent\textbf{ACTIVA successfully estimates the causal shift distributions.}
 Table~\ref{tab:main_results} shows first that ACTIVA consistently improves over the Gaussian Conditional baseline on NLL and, most importantly, strongly reduces spurious effects on non-descendants. This is the clearest evidence that ACTIVA learns predominantly causal rather than merely correlational structure: variables whose true causal effect is zero are assigned substantially smaller distribution shifts than under the conditional Gaussian baseline.

% \noindent\textbf{Near perfect estimation in 2 variable datasets.}
On the two-variable datasets, ACTIVA achieves very low errors overall. The strongest result appears on \emph{Beta 2}, where ACTIVA outperforms both MACE-TNP and the Linear Causal baseline, which has privileged information, across the reported metrics, consistent with the advantage of a flexible mixture decoder in a non-Gaussian setting. On \emph{Gauss 2}, ACTIVA remains close to the asymptotic oracle Linear Causal baseline. Furthermore, ACTIVA improves over MACE-TNP on \emph{Gauss 2}, although not uniformly across every condition and metric.

 On the larger synthetic datasets \emph{Gauss 8} and \emph{Beta 8}, the results reveal complementary strengths across amortized methods. ACTIVA remains clearly superior to the correlational baseline, but MACE-TNP attains lower energy distance and lower NLL on several 8-variable synthetic subsets. This suggests that, for larger synthetic graphs, MACE-TNP can better match sample-space geometry, whereas ACTIVA is still competitive but no longer uniformly best.

% \noindent\textbf{Accurate predictions on biologically plausible data.}
The semi-synthetic SERGIO results highlight a different strength of ACTIVA. Here, ACTIVA achieves the best NLL in all reported marginals and remains competitive in energy distance, indicating particularly strong density estimation on biologically plausible data. This holds against both graph-privileged baselines: lifting the linear-Gaussian restriction does not yield a stronger competitor, as the Nonlinear Causal baseline reaches an NLL of 14.83 in-distribution and 20.31 out-of-distribution, against 14.07 and 18.55 for the linear one, without improving on it in energy distance either. The comparison is informative about what the graph provides. Given the true graph, the Nonlinear Causal baseline is accurate on the first moment: its mean absolute ATE error on non-descendants, where the true effect is exactly zero, is 0.176, and it is slightly ahead of ACTIVA in non-descendant energy distance (0.79 against 0.87). Its NLL on those same coordinates is nonetheless 13.33 against 8.99, and on descendants 12.15 against 3.94. What the graph buys is therefore the moment rather than the density: modularity is exact once the graph is known, but the conditional mechanisms and their noise densities must still be estimated from the 30 observational samples available per task, which is precisely the regime amortization targets. 

The OOD row is a simulator-mismatch study:  ACTIVA is applied to it without retraining, attaining ERG 2.96 and NLL 17.97 against 8.22 and 90.87 for the correlational baseline and 1.07 and 18.55 for the graph-privileged linear one. Because the gene count changes, this also exercises intervention targets never seen in training, which permutation equivariance makes representable. Performance still degrades relative to the in-distribution setting, most notably on descendant energy distance, and the shift is parametric within a fixed generative family, so we read the result as robustness to simulator misspecification of that kind rather than as uniformly strong generalization.

% \noindent\textbf{Trade-offs regarding MACE-TNP.}
Comparing ACTIVA and MACE-TNP reveals complementary trade-offs. ACTIVA tends to provide better-calibrated joint densities when the intervened variables are included in the target distribution, as reflected by its stronger NLL in the \textit{All} condition. In contrast, MACE-TNP is often stronger in energy distance on the larger synthetic problems when the target variable is not of interest. Taken together, these results suggest that ACTIVA is particularly attractive when the goal is to estimate the full post-intervention density from observational data and a weakly specified intervention query, rather than only a marginal over non-intervened variables.

Overall, the results in Table~1 show that ACTIVA successfully leverages causal information at inference time. In a single forward pass, it produces competitive interventional distributions, substantially improves over a correlational baseline, and remains especially strong on small graphs and semi-synthetic data, given only observational data and a weakly specified intervention query.

%\begin{table}[b]
%\centering

%\caption{Average inference performance of ACTIVA trained on semi-syntetic data and tested on a held-out test set and an out-of-distribution set.} % These results are on the test set of the SERGIO training data and on the test set of the ood data
%\begin{tabular}{@{}lcccccc@{}}
%\toprule
%       & \multicolumn{3}{c}{ACTIVA} & \multicolumn{3}{c}{Baseline} \\ 
%Data                      & MMD    & WSD     & ERG     & MMD     & WSD      & ERG     \\ \midrule
%SERGIO                    & 0.34   & 4.3    & 1.7    & 0.50    & 3.9     & 2.8    \\
%OOD                   & 0.37   & 7.6    & 3.7    & 0.61    & 8.0     & 8.0   \\
% \bottomrule
%\end{tabular}

%\label{tb:SERGIO}
%\end{table}

% If you truly want it to always fit, uncomment the resizebox line and its closing `%}` and
% remove the `\scriptsize`/`\tabcolsep` tweaks. That will automatically scale the table to the page width.

\subsection{Downstream Task Performance}\label{sec:downstream}
\begin{figure}[ht]
    \centering
        \includegraphics[width=\linewidth]{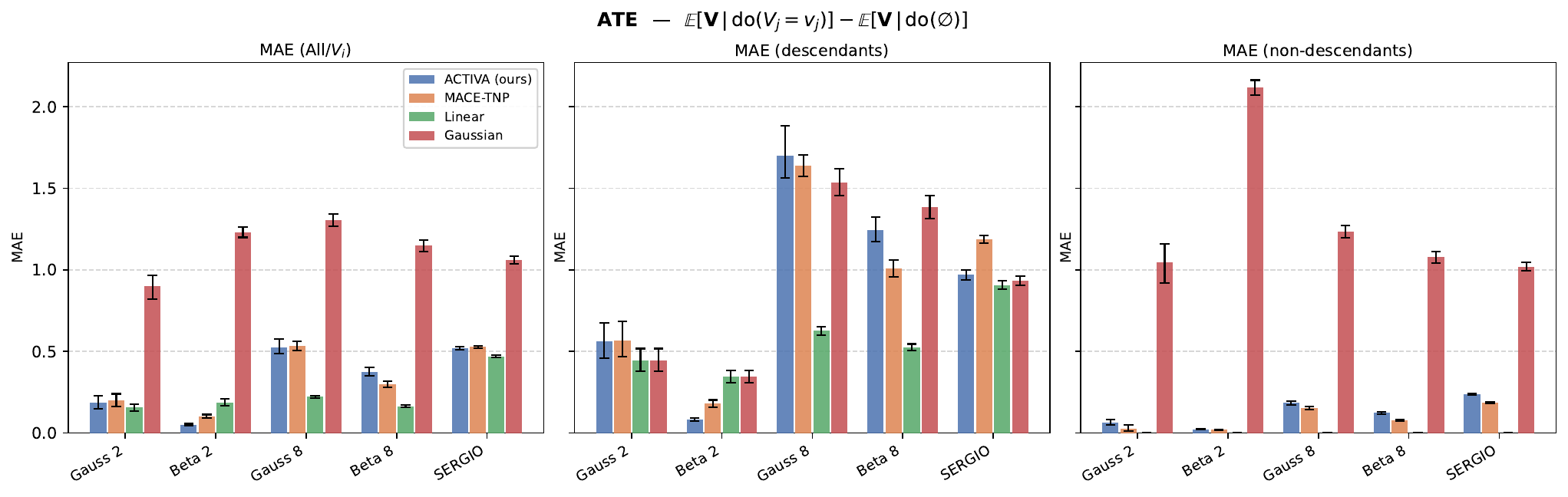}
        \caption{Mean absolute error of the ATE of the predicted distributions w.r.t. the ground-truth sample ATE. The left panel shows the error on the full joint distribution (excluding the intervened variable), the middle panel shows the error for the causal descendants, and the right panel for the causal non-descendants of the intervention target. The different approaches and baselines are color-coded. Lower is better.}
        \label{fig:ate}
\end{figure}

We finally assess whether the distributional predictions of ACTIVA translate into accurate downstream effect estimates. This is especially of interest, as we are interested in the ability to use our methods to derive causal conclusions; even when distributional aspects like covariances and multi-modality are not of interest. To this end, Figure~\ref{fig:ate} reports the mean absolute error of the ATE for the full set of non-intervened variables, for descendants only, and for non-descendants only.  

% \noindent\textbf{Capturing mostly causal information.}
Figure~\ref{fig:ate} suggests the same qualitative pattern as Table~\ref{tab:main_results}. ACTIVA clearly improves over the Gaussian Conditional baseline, especially on non-descendants, indicating that it captures not only causal effects but also the absence of effects. Thus, the gains in Section~7.2 are not limited to density modeling alone; they carry over to a standard causal estimand.

% \noindent\textbf{State-of-the-art average effect prediction.}
Relative to MACE-TNP, ACTIVA is broadly competitive rather than uniformly superior. The clearest advantage appears on \emph{Beta 2}, whereas MACE-TNP is stronger on \emph{Beta 8}, and the remaining datasets are comparatively close. This suggests that the rich distributional modeling of ACTIVA does not come at the expense of average-effect estimation, but it also does not automatically translate into uniformly better ATE performance compared to MACE-TNP.

\noindent\textbf{Competitive with graph-privileged baselines on small graphs and semi-synthetic data.}
ACTIVA is also competitive with the privileged-information Linear Causal baseline on the smaller problems and on SERGIO. We use oracle terminology only for the Gaussian datasets, where that baseline is correctly specified; on Beta and even more so SERGIO datasets it is a privileged-information comparison. Especially for the \emph{Beta 2} dataset, these results indicate that predicting the post-interventional distribution as a GMM helps with the downstream mean prediction. However, for larger graphs, this advantage over simple Gaussian prediction is quickly offset when the underlying causal graph is known.

These results are consistent with ACTIVA exploiting causal rather than merely correlational structures from observational data and weakly specified intervention queries. Furthermore, this benefit remains visible even when the final goal is a standard treatment-effect estimate rather than the full interventional distribution.

\subsection{Empirical Evaluation of Theoretical Results}\label{sec:non_identifiability}

Sections~\ref{sec:qualitative_results}--\ref{sec:downstream} give an account on the quality of the predicted post-interventional distribution. In this Section, we empirically validate Proposition 1 and 2 directly. Specifically, we test whether when
$[\mathcal{M}^o]$ contains several models, the objective targets their mixture, weighted by
the simulator distribution $p_{tr}(\mathcal{M}\mid[\mathcal{M}^o], do(V))$. Because the synthetic datasets in Section~\ref{sec:data} are identifiable from observational data, this is the only experiment designed to illustrate the behavior in a non-identifiable setting.

\noindent\textbf{Setup.}
For this experiment, we depart from Section~\ref{sec:experimental_setup} in a few ways. We generate a dataset from two fixed SCMs:
\begin{equation}\label{eq:ni_scms}
\mathcal{M}_A:\;
\begin{cases}
X = \varepsilon_X, & \varepsilon_X \sim \mathcal{N}(0,1)\\
Y = 0.8 X + \varepsilon_Y, & \varepsilon_Y \sim \mathcal{N}(0,\,1-0.8^2)
\end{cases}
\qquad
\mathcal{M}_B:\;
\begin{cases}
Y = \eta_Y, & \eta_Y \sim \mathcal{N}(0,1)\\
X = 0.8 Y + \eta_X, & \eta_X \sim \mathcal{N}(0,\,1-0.8^2),
\end{cases}
\end{equation}
with mutually independent noise. Choosing the edge weight and noise variance in this
way makes the pair exactly self-mirroring: both models induce the same observational distribution
$\mathcal{N}\!\bigl(\boldsymbol{0},\left[\begin{smallmatrix}1&0.8\\0.8&1\end{smallmatrix}\right]\bigr)$,
so $[\mathcal{M}^o]=\{\mathcal{M}_A,\mathcal{M}_B\}$ however much observational data is
supplied. Their interventional distributions, however, differ: under $do(X\!=\!5)$,
$p_{\mathcal{M}_A}(Y\mid do(X\!=\!5))=\mathcal{N}(4.00,\,0.36)$ whereas
$p_{\mathcal{M}_B}(Y\mid do(X\!=\!5))=\mathcal{N}(0,\,1)$.% This is the \emph{Gauss 2} setting reduced to its sharpest form; the equivalence holds precisely \emph{because} the noise is Gaussian, which is also what makes the closed-form target available.}

\begin{figure}[t]
\centering
\begin{subfigure}[t]{0.48\linewidth}
    \centering
    \includegraphics[width=\linewidth]{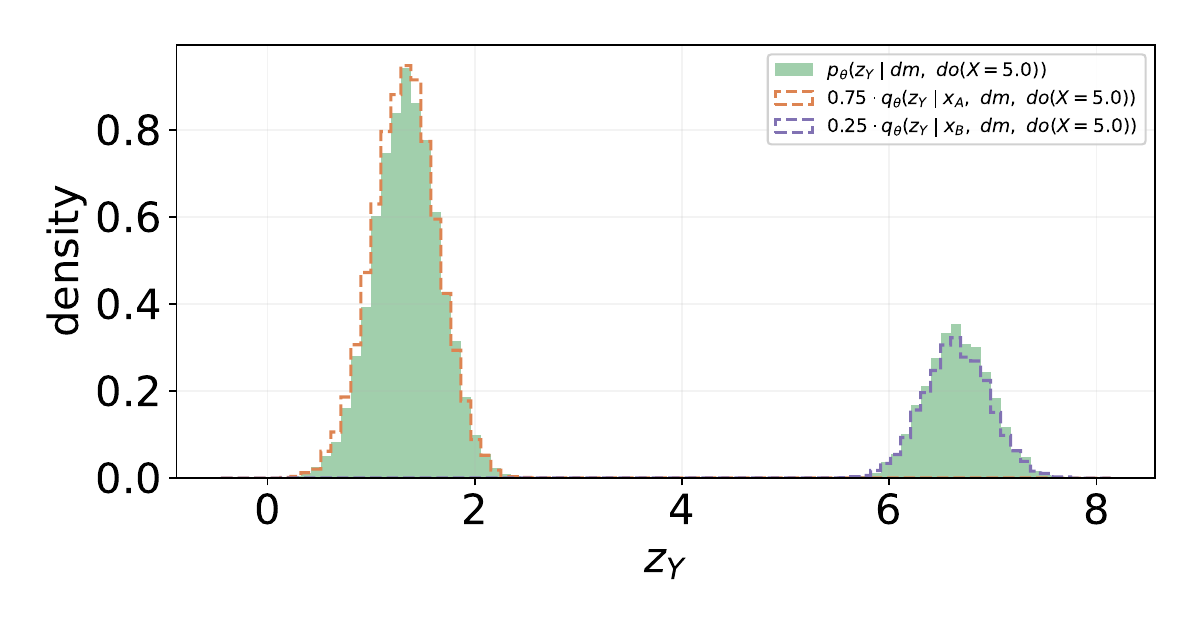}
    \caption{Latent space, $z_Y$ coordinate.}
    \label{fig:ni_latent}
\end{subfigure}
\hfill
\begin{subfigure}[t]{0.48\linewidth}
    \centering
    \includegraphics[width=\linewidth]{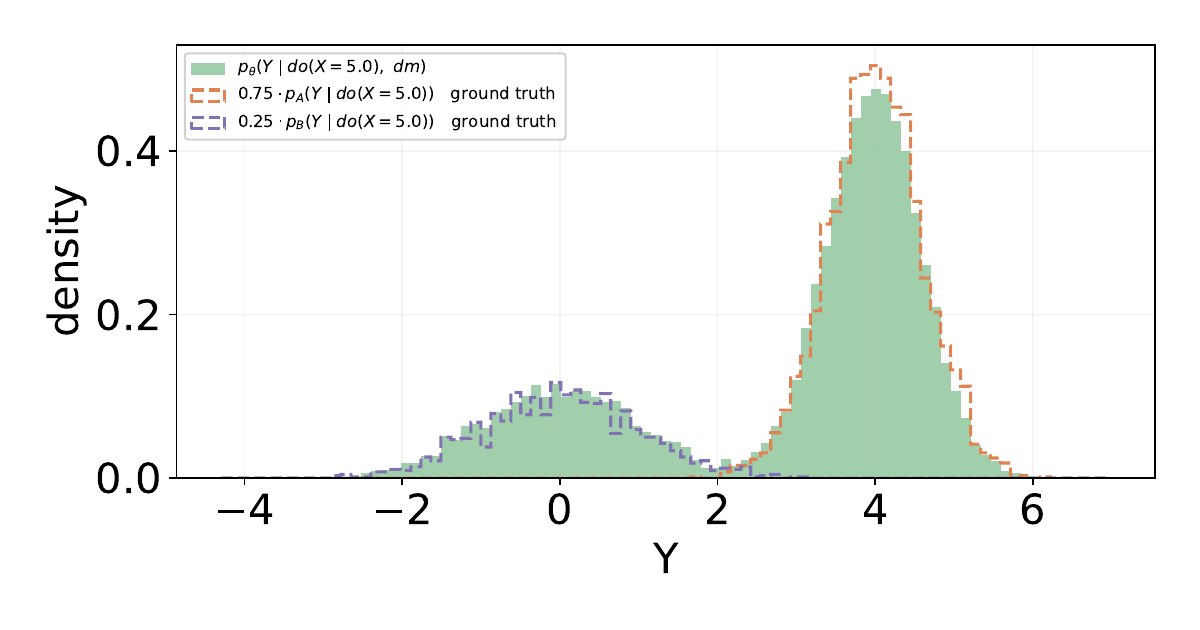}
    \caption{Post-interventional distribution of $Y$.}
    \label{fig:ni_prior}
\end{subfigure}
\caption{Behavior under provable non-identifiability, given only the shared
observational input and the query $do(X\!=\!5)$. \textbf{(a)} Samples from the full $C=20$
component conditional prior (green). Dashed
curves (orange and blue) are the two posteriors scaled by their branch rate, so they sum to the aggregated
posterior. \textbf{(b)} Samples from the decoded prior (green) and ground-truth interventional samples scaled by their simulator weight, so they sum to the target in
\Eqref{eq:ni_mixture} (dashed, orange and blue).}
\label{fig:ni}
\end{figure}

To avoid finite-sample effects to make the two models distinguishable by their observational data, all
training rows share a single observational input $\boldsymbol{D}^{\mathcal{M}^o}$ of $50$
samples and the same intervention representation $\boldsymbol{I}_{i,t}$; only the
interventional samples differ between tasks. A fraction $\pi_A=0.75$ of the $1024$ task draws
them from $\mathcal{M}_A$ and the rest from $\mathcal{M}_B$, so that
$p_{tr}(\mathcal{M}_A\mid[\mathcal{M}^o], do(V))=0.75$ by construction and Proposition~2
predicts exactly
\begin{equation}\label{eq:ni_mixture}
p_{\theta^*}\!\left(Y \mid \boldsymbol{D}^{\mathcal{M}^o}, do(X\!=\!5)\right)
= 0.75\,\mathcal{N}(4.00,\,0.36) \;+\; 0.25\,\mathcal{N}(0,\,1).
\end{equation}
%\added{Because the prior sees identical inputs on every row, no encoder capacity can separate the branches. We choose $\pi_A$ asymmetric deliberately: the pair in \Eqref{eq:ni_scms} is self-mirroring, so under $\pi_A=0.5$ any model with a symmetric inductive bias, or one that simply fails to break a tie, would also produce equal weights; an asymmetric proportion separates recovering the simulator weights from splitting evenly by symmetry. 
Since the target is known in closed form, we compare against it directly rather than against the baselines and sample-based metrics of
Sections~\ref{sec:baselines}--\ref{sec:marginal_subsets}. We split the data $80/10/10$
into train, validation and test with the branch ratio preserved in each split We train from scratch with hidden
and embedding dimension $16$ and two encoder and two decoder blocks,
retaining the $C=20$ prior components and $B=2$ decoder components used throughout the paper. Note that we use ten times more prior components than the equivalence class has members, so that
concentrating onto two modes remains an empirical outcome rather than an architectural
constraint. We train the ACTIVA model under otherwise identical conditions to the previous experiments.% Appendix~\ref{apx:non_identifiability} gives the construction, the remaininghyperparameters, and the evaluation protocol.

\noindent\textbf{The prior matches the aggregated posterior.}
Figure~\ref{fig:ni}~(a) shows the latent view. Posterior means form two tight clusters ($z_Y = 1.325$ and $6.646$, separated by 16.8 latent component standard deviations). The prior's occupied components sit on those clusters, matching the cluster means to within 0.018. Taken as whole distributions the prior reproduces the aggregate posterior's first two moments: mean 2.767 vs 2.655, variance 5.639 vs 5.410. %We attribute most of the residual (+0.11, +0.23) to the weights being slightly off target (Result 2) due to SGD; we make no claim of exact equality.
 Of the 20 available components the prior uses six carrying mass above $10^{-4}$,  two carrying 91\% of the mass, the remaining four also sitting on a cluster centre (all within 0.5 component standard deviations) and fourteen driven below $10^{-4}$. Given the freedom to smear across latent space or to interpolate between the branches, the model did neither.

\noindent\textbf{The estimation is the weighted mixture.}
Decoding the prior by sampling $50{,}000$ latents, rather than the $10$ used in Section~\ref{sec:training_and_inference}, yields the bimodal predictive in Figure~\ref{fig:ni}~(b). The larger sample serves to make the Monte-Carlo error on the reported quantities negligible, which is affordable at this problem size; it is not required for the effect itself. The prior-path predictive decomposes into two modes at 3.993 and $-0.006$ with variances 0.363 and 1.016, against ground truth 4.000 / 0.000 and 0.360 / 1.000. The mixture weight, read directly off the conditional prior's mixture logits, summed over the components assigned to each cluster, is 0.732 against the simulator's $\pi_A = 0.750$. The prior tracks the rate at which each SCM occurs in the training distribution rather than splitting mass by symmetry, exactly the estimand Proposition~2 characterises.
Separating the two paths, of the 50,000 individually decoded mixtures, 73.1\% place essentially all their mass on $\mathcal{M}_A$ and 26.9\% on $\mathcal{M}_B$; none is bimodal. Each decoded output commits to a single model, and the mixture arises because the prior scatters latents across two separated regions. Exactly the mechanism the theory predicts.

% \noindent\textbf{\added[id=R]{Model uncertainty is carried by the prior.}}
% \added[id=R]{The formulation assigns two sources of uncertainty to different components. Uncertainty over which model generated the data resides in the prior over latents (Proposition~1), whereas variation within a model, including the value distribution behind an intervention label and exogenous noise, is carried by the decoder given a latent sample. The predictive variance accordingly splits into a between part, attributable to model uncertainty, and a within part, attributable to within-model variation. Since every decoded output commits to a single model, the between part is here the variance between the two modes, $\rho_A\rho_B(\mu_A-\mu_B)^2$ with mode masses $\rho$ and means $\mu$, and the within part is the mass-weighted mode variance. The values reported above give $3.14$ between and $0.54$ within, a total of $3.68$, against $3.00$, $0.52$ and $3.52$ for the target in \Eqref{eq:ni_mixture}. Model uncertainty thus accounts for about $85\%$ of the predictive variance and is recovered by the prior rather than absorbed by the decoder. Two caveats apply. The separation is exact only in the idealized limit of Section~\ref{sec:theory}: with finite samples, model uncertainty that the encoder cannot resolve may remain in the decoder. Moreover, within-label heterogeneity and exogenous noise both reside in the decoder, so the formulation does not separate these two from each other.}

The findings line up with the theory point for point. That the prior reproduces the aggregated posterior, is Proposition~1 realized in a finite-capacity model. That decoding it returns the two post-interventional distributions, at the right locations and widths, is the mixture structure of
Proposition~2. That the weigths reproduce $\pi_A$ identifies the mixture weights as the conditional simulator distribution $p_{tr}(\mathcal{M}\mid[\mathcal{M}^o], do(V))$, which is
the part of Proposition~2 that distinguishes it from a generic statement about
multi-modality: ACTIVA does not merely hedge, it hedges at the rate the training
distribution prescribes. 
In this experiment, the model-space structure is carried by the prior mixture, which suggests using at least $|[\mathcal{M}^o]|$ prior components. This is, however, not a strict requirement since a sufficiently expressive decoder can also produce a multimodal prediction from a unimodal prior.

\section{Related Work}
\label{sec:related_work}

Deep learning approaches to causal inference have advanced the modeling of observational, interventional, and counterfactual distributions, yet, most still require either known or learnable causal graphs, relatively strong structural assumptions, or dataset-specific fitting at test time. This holds for VAE-based models~\citep{Yang2021CausalVAE:Models,Qi2023CMVAE:Meta-Learning}, diffusion generative models~\citep{Sanchez2022DiffusionEstimation,Chao2023InterventionalModels}, graph-neural and flow-based approaches~\citep{Zecevi2021RelatingModels,Sanchez-Martin2022VACA:Queries,Poinsot2024LearningChallenges}, and adversarial network approaches~\citep{Rahman2024ModularInference}.

More recently, a line of work has begun to amortize causal inference across datasets. Foundation-model and meta-learning approaches  train a single model across many simulated tasks and then perform zero-shot or in-context inference on new datasets~\citep{Balazadeh2025CausalPFN:Learning,Robertson2025DO-PFN:ESTIMATION,Bynum2025BlackPrediction,Ma2025FoundationNetworks}. These methods substantially reduce per-dataset computation, but they primarily target causal-effect summaries or conditional interventional quantities in identifiable settings, rather than the full joint post-intervention distribution over all variables.

The closest recent work to ours is MACE-TNP, which meta-learns interventional distributions directly from observational data under causal-graph uncertainty~\citep{Dhir2025EstimatingMeta-Learning}. Relative to MACE-TNP, ACTIVA differs in two main ways. First, ACTIVA is designed to predict the full joint post-intervention distribution rather than the interventional distribution of a designated outcome variable. Second, ACTIVA is tailored to weakly specified interventions, conditioning on intervention identifiers and targets rather than requiring a fully specified intervention value at inference time. 

Our work is also related to approaches that amortize inference over SCMs, including zero-shot and meta-learned SCM inference methods~\citep{Lowe2022AmortizedData,Nilforoshan2023Zero-shotLearning,Sauter2024CORE:Learning,Mahajan2024Zero-ShotModels}. These methods aim to recover or simulate causal models themselves, whereas our goal is different: we seek direct estimation of post-intervention distributions from observational data and an intervention query without requiring explicit graph recovery at inference time.

Finally, some methods focus on specialized causal-estimation settings, such as treatment-outcome estimation, limited overlap, or marginal interventional supervision, without targeting the full joint interventional distribution across all variables~\citep{Louizos2017CausalModels,Vowels2021TargetedInference,Wu2023-Intact-VAE:Overlap,Vanderschueren2023NOFLITE:Distributions,Garrido2024EstimatingData}. In contrast, ACTIVA is aimed at the amortized estimation of the full post-interventional distribution from observational inputs and weakly specified intervention queries, while explicitly representing ambiguity through a mixture distribution induced by observationally compatible causal models.

%\subsection{Remarks}
%There is a whole body of work that focuses on learning interventional, observational, counterfactual distributions for each variable separately (see \citep{Javaloy2023CausalPractice} introduction. "e.g., normalizing flows (NFs) [23, 26, 27], generative adversarial networks (GANs) [20, 39], variational autoencoders (VAEs) [15, 40], Gaussian processes (GPs) [15], or denoising diffusion probabilistic models (DDPMs) [2]—to iteratively estimate the conditional distribution of each observed variable given its causal parents, thus using an independent DNN per observed variable", but they suffer from error propagation and having to estimates many parameters \citep{Javaloy2023CausalPractice}

\section{Conclusion}\label{sec:conclusion}
We introduced ACTIVA, a conditional variational autoencoder for amortized estimation of post-intervention distributions from observational data and weakly specified intervention queries. Unlike classical causal-effect estimators that return only summary quantities, ACTIVA is designed to predict the full joint distribution after intervention, thereby capturing both downstream effect estimates and richer forms of causal uncertainty, such as multi-modality. Our theoretical analysis provides a population-level characterization of ACTIVA's estimand: under idealized conditions, ACTIVA's learning objective targets a mixture over the interventional distributions of models that remain observationally compatible with the input, making explicit how observational ambiguity is reflected in the estimand.

Empirically, ACTIVA demonstrates that this theory-guided amortized approach is effective in finite models. Across synthetic datasets and biologically realistic gene-expression simulations, ACTIVA substantially outperforms a purely correlational baseline, reduces spurious effects on non-descendants, and achieves competitive performance relative to strong amortized baselines. Its distributional predictions also translate into competitive downstream ATE estimates, showing that the learned post-interventional distributions capture causal structure in a way that remains useful for standard causal-effect summaries. On a pair of observationally equivalent models sharing a single observational dataset, the learned prior concentrates onto the two members of the equivalence class and reproduces the rate at which each occurs in the training distribution, showing that the mixture structure predicted by our analysis is attained by the finite-capacity model and not only in the population limit.
Taken together, these findings position ACTIVA as a promising approach when the goal is estimation of full post-interventional distributions from observational data and partially specified interventions, without per-task fitting or explicit causal model recovery at inference time. These findings concern causal models drawn from the training distribution; beyond it, only a parametric shift of the simulator was tested.

More broadly, we view this work as a step toward amortized causal inference systems that represent model ambiguity explicitly in the predicted post-intervention distribution, rather than collapsing it into a single point estimate.
%do not hide structural ambiguity behind a single point estimate but instead represent it directly in the predicted post-intervention distribution. 
Important directions for future work include further tightening the connection between the population-level theory and finite-sample implementations beyond the controlled setting studied here, improving performance on larger and more weakly matched out-of-distribution settings, characterizing how sensitive the prediction is to a mismatch between $p_{tr}$ and the deployment distribution,  including the untested regime in which one intervention label spans widely separated values, and evaluating the approach on real paired observational and perturbation data, where weakly specified interventions arise naturally and which the present simulator-based evaluation does not provide. A further direction is implementing the extension to partial causal knowledge outlined in Section~\ref{sec:architecture}, for which our analysis predicts that the mixture collapses whenever the remaining compatible models agree on the query.%Since such constraints shrink the observational equivalence class, our analysis predicts that the mixture in Proposition~2 then has fewer members and collapses whenever the remaining ones agree on the queried distribution (Corollary~1).

%\section*{Impact Statement}
%This paper presents work whose goal is to advance the field of Machine Learning. There are many potential societal consequences of our work, none of which we feel must be specifically highlighted here.

%\subsubsection*{Author Contributions}
%If you'd like to, you may include  a section for author contributions as is done in many journals. This is optional and at the discretion of the authors.

% \section*{Reproducibility Statement} %TODO discuss
% We took several steps to support reproducibility. An anonymous repository with training and evaluation code, and instructions is provided in the supplementary materials (see the code link in the main text). Our model details and architectural choices are specified in Section \ref{sec:architecture}, with the theoretical assumptions and proofs discussed in detail in Section \ref{sec:theory}. Complete dataset generation procedures (synthetic and SERGIO), splits (including OOD), and intervention settings are documented in Appendix \ref{apx:datasets}; evaluation metrics and their implementations are described in Appendix \ref{apx:metrics}; and all optimization settings, hyperparameters, seeds, and hardware are summarized in Appendix \ref{apx:hyperparameters}. We also describe baselines and inference procedures within Section \ref{sec:experiments} to facilitate like-for-like comparisons. Lastly, we provide the datasets needed to reproduce some of our experiments.

\section*{Acknowledgments}
We thank Emile van Krieken, Majid Mohammadi and Álvaro Serra-Gómez, and the anonymous reviewers for their helpful inputs and feedback to this project.

This research was partially funded by the Hybrid Intelligence Center, a 10-year programme funded by the Dutch Ministry of Education, Culture and Science through the Netherlands Organisation for Scientific Research, https://hybrid-intelligence-centre.nl, grant number 024.004.022

% \subsubsection*{Reproducability Statement}

\bibliography{references}
\bibliographystyle{tmlr}
\onecolumn

\appendix
% \section{Appendix}

\section{Derivation of KL Decomposition}\label{apx:kl_decomposition}
In this section, we derive the KL decomposition used in Section~\ref{sec:theory}. Since the theory is stated at the level of the joint simulator distribution $p_{tr}(\mathcal{M}, do(V))$ over causal models and intervention queries, we work directly with the corresponding population objective. The finite-data objective used in practice can be viewed as a Monte Carlo approximation of the expectations below. The weight $\beta$ of \Eqref{eq:activa_elbo} is omitted throughout: it multiplies the entire KL term by a positive constant, which leaves the minimizer in $\eta$ unchanged, so the decomposition below and Proposition~1 hold for every $\beta>0$.

Before restricting to a fixed observational equivalence class and query, note that the KL contribution to the population objective decomposes by the law of total expectation over pairs $S=([\mathcal{M}],do(V))$:
\begin{equation}\label{eq:appendix_total_expectation}
\mathbb{E}_{\mathcal{\mathbf{D}}^{tr}\sim p_{tr}(\mathbf{D}^{tr})}
\Big[
\operatorname{KL}\big(
q_\phi(\boldsymbol{z}\mid \mathcal{\mathbf{D}}^{tr})
\,\|\, 
p_\eta(\boldsymbol{z}\mid \boldsymbol{D}^{\mathcal{M}}, do(V))
\big)
\Big]
=
\mathbb{E}_{S\sim p_{tr}(S)}
\Big[
\mathbb{E}_{\mathcal{\mathbf{D}}^{tr}\sim p_{tr}(\mathbf{D}^{tr}\mid S)}
\big[
\operatorname{KL}(\cdots)
\big]
\Big].
\end{equation}
Hence, it is sufficient to analyze the inner expectation for an arbitrary fixed pair $S$. Since the outer expectation in \Eqref{eq:appendix_total_expectation} is only a weighted average over such terms, the dependence of the full objective on the prior parameters $\eta$ is completely determined by the class-conditional decomposition derived below.

Fix an observational equivalence class $[\mathcal{M}^o]$ and a query intervention $do(V)$. Restricting the expected KL term in \Eqref{eq:amortized_objective} to tasks whose observational component lies in $[\mathcal{M}^o]$ and whose query equals $do(V)$ gives
\begin{equation}\label{eq:appendix_conditional_kl}
    \mathbb{E}_{\mathcal{\mathbf{D}}^{tr}\sim p_{tr}(\mathbf{D}^{tr}\mid [\mathcal{M}^o], do(V))}
    \Big[
        \operatorname{KL}\big(
            q_\phi(\boldsymbol{z}\mid \mathcal{\mathbf{D}}^{tr})
            \,\|\, 
            p_\eta(\boldsymbol{z}\mid \boldsymbol{D}^{\mathcal{M}^o}, do(V))
        \big)
    \Big].
\end{equation}
Following the aggregated-posterior decomposition for amortized variational objectives of \citet{Hoffman2016ELBOBound}, we rewrite this expected KL term by introducing the corresponding aggregated posterior over tasks in the same observational equivalence class and intervention query.

Expanding the KL divergence yields
\begin{align}
    &\mathbb{E}_{\mathcal{\mathbf{D}}^{tr}\sim p_{tr}(\mathbf{D}^{tr}\mid [\mathcal{M}^o], do(V))}
    \Big[
        \operatorname{KL}\big(
            q_\phi(\boldsymbol{z}\mid \mathcal{\mathbf{D}}^{tr})
            \,\|\, 
            p_\eta(\boldsymbol{z}\mid \boldsymbol{D}^{\mathcal{M}^o}, do(V))
        \big)
    \Big] \notag \\
    &= 
    \mathbb{E}_{\mathcal{\mathbf{D}}^{tr}\sim p_{tr}(\mathbf{D}^{tr}\mid [\mathcal{M}^o], do(V))}
    \left[
        \int
        q_\phi(\boldsymbol{z}\mid \mathcal{\mathbf{D}}^{tr})
        \log
        \frac{
            q_\phi(\boldsymbol{z}\mid \mathcal{\mathbf{D}}^{tr})
        }{
            p_\eta(\boldsymbol{z}\mid \boldsymbol{D}^{\mathcal{M}^o}, do(V))
        }
        \, d\boldsymbol{z}
    \right].
\end{align}
Writing the outer expectation as an integral over tasks gives
\begin{align}
    &=
    \int
    \left[
        \int
        q_\phi(\boldsymbol{z}\mid \mathcal{\mathbf{D}}^{tr})
        \log
        \frac{
            q_\phi(\boldsymbol{z}\mid \mathcal{\mathbf{D}}^{tr})
        }{
            p_\eta(\boldsymbol{z}\mid \boldsymbol{D}^{\mathcal{M}^o}, do(V))
        }
        \, d\boldsymbol{z}
    \right]
    p_{tr}(\mathcal{\mathbf{D}}^{tr}\mid [\mathcal{M}^o], do(V))
    \, d\mathcal{\mathbf{D}}^{tr} \notag \\
    &=
    \int
    p_{tr}(\mathcal{\mathbf{D}}^{tr}\mid [\mathcal{M}^o], do(V))
    q_\phi(\boldsymbol{z}\mid \mathcal{\mathbf{D}}^{tr})
    \log
    \frac{
        q_\phi(\boldsymbol{z}\mid \mathcal{\mathbf{D}}^{tr})
    }{
        p_\eta(\boldsymbol{z}\mid \boldsymbol{D}^{\mathcal{M}^o}, do(V))
    }
    \, d\boldsymbol{z}\, d\mathcal{\mathbf{D}}^{tr}.
\end{align}
Defining the joint distribution
\begin{equation}
    q_\phi(\mathcal{\mathbf{D}}^{tr}, \boldsymbol{z}\mid [\mathcal{M}^o], do(V))
    :=
    p_{tr}(\mathcal{\mathbf{D}}^{tr}\mid [\mathcal{M}^o], do(V))\,
    q_\phi(\boldsymbol{z}\mid \mathcal{\mathbf{D}}^{tr}),
\end{equation}
we can rewrite the expected KL term as
\begin{equation}
    \int
    q_\phi(\mathcal{\mathbf{D}}^{tr}, \boldsymbol{z}\mid [\mathcal{M}^o], do(V))
    \log
    \frac{
        q_\phi(\boldsymbol{z}\mid \mathcal{\mathbf{D}}^{tr})
    }{
        p_\eta(\boldsymbol{z}\mid \boldsymbol{D}^{\mathcal{M}^o}, do(V))
    }
    \, d\boldsymbol{z}\, d\mathcal{\mathbf{D}}^{tr}.
\end{equation}

We define the aggregated posterior
\begin{equation}\label{eq:appendix_agg_posterior}
    \bar q_\phi(\boldsymbol{z}\mid [\mathcal{M}^o], do(V))
    :=
    \int
    q_\phi(\boldsymbol{z}\mid \mathcal{\mathbf{D}}^{tr})\,
    p_{tr}(\mathcal{\mathbf{D}}^{tr}\mid [\mathcal{M}^o], do(V))
    \, d\mathcal{\mathbf{D}}^{tr}.
\end{equation}
Adding and subtracting $\log \bar q_\phi(\boldsymbol{z}\mid [\mathcal{M}^o], do(V))$ inside the logarithm yields
\begin{align}
    &\int
    q_\phi(\mathcal{\mathbf{D}}^{tr}, \boldsymbol{z}\mid [\mathcal{M}^o], do(V))
    \log
    \frac{
        q_\phi(\boldsymbol{z}\mid \mathcal{\mathbf{D}}^{tr})
    }{
        p_\eta(\boldsymbol{z}\mid \boldsymbol{D}^{\mathcal{M}^o}, do(V))
    }
    \, d\boldsymbol{z}\, d\mathcal{\mathbf{D}}^{tr} \notag \\
    &= \int
    q_\phi(\mathcal{\mathbf{D}}^{tr}, \boldsymbol{z}\mid [\mathcal{M}^o], do(V))
    \log
    \frac{
        q_\phi(\boldsymbol{z}\mid \mathcal{\mathbf{D}}^{tr})
    }{
        \bar q_\phi(\boldsymbol{z}\mid [\mathcal{M}^o], do(V))
    }
    \, d\boldsymbol{z}\, d\mathcal{\mathbf{D}}^{tr} \notag \\
    &\quad +
    \int
    q_\phi(\mathcal{\mathbf{D}}^{tr}, \boldsymbol{z}\mid [\mathcal{M}^o], do(V))
    \log
    \frac{
        \bar q_\phi(\boldsymbol{z}\mid [\mathcal{M}^o], do(V))
    }{
        p_\eta(\boldsymbol{z}\mid \boldsymbol{D}^{\mathcal{M}^o}, do(V))
    }
    \, d\boldsymbol{z}\, d\mathcal{\mathbf{D}}^{tr}.
\end{align}
The first term is the conditional mutual information between the task and the latent variable under $q_\phi$,
\begin{equation}
    \mathbb{I}_{q_\phi}\!\left(
        \mathcal{\mathbf{D}}^{tr}; \boldsymbol{z}
        \mid [\mathcal{M}^o], do(V)
    \right),
\end{equation}
while in the second term the dependence on $\mathcal{\mathbf{D}}^{tr}$ disappears inside the logarithm. We can therefore marginalize out $\mathcal{\mathbf{D}}^{tr}$ using the definition of $\bar q_\phi$ in \Eqref{eq:appendix_agg_posterior}, which gives
\begin{align}\label{eq:KL_decomposed_extended_updated}
    &\mathbb{E}_{\mathcal{\mathbf{D}}^{tr}\sim p_{tr}(\mathbf{D}^{tr}\mid [\mathcal{M}^o], do(V))}
    \Big[
        \operatorname{KL}\big(
            q_\phi(\boldsymbol{z}\mid \mathcal{\mathbf{D}}^{tr})
            \,\|\, 
            p_\eta(\boldsymbol{z}\mid \boldsymbol{D}^{\mathcal{M}^o}, do(V))
        \big)
    \Big] \notag \\
    &= 
    \operatorname{KL}\!\left(
        \bar q_\phi(\boldsymbol{z}\mid [\mathcal{M}^o], do(V))
        \,\middle\|\,
        p_\eta(\boldsymbol{z}\mid \boldsymbol{D}^{\mathcal{M}^o}, do(V))
    \right)
    +
    \mathbb{I}_{q_\phi}\!\left(
        \mathcal{\mathbf{D}}^{tr}; \boldsymbol{z}
        \mid [\mathcal{M}^o], do(V)
    \right).
\end{align}

In the main text, we focus on the KL term in \Eqref{eq:KL_decomposed_extended_updated}, since it is the only term that depends on the prior parameters $\eta$. This yields the expression used in \Eqref{eq:KL_decomposed_population_integral}.

\section{Proposition 2 Proof}\label{apx:prop2_proof}
Here we give the full proof of Proposition~2 from the main text. We start by re-stating the proposition.

\textbf{Proposition 2.} Under Assumptions 1--3, the ACTIVA predictive distribution for an input $\bigl(\boldsymbol{D}^{\mathcal{M}^o}, do(V)\bigr)$ is given by
\begin{equation}
    p_{\theta^*}(\boldsymbol{V}\mid \boldsymbol{D}^{\mathcal{M}^o}, do(V))
    =
    \int
    p_{\mathcal{M}}(\boldsymbol{V}\mid do(V))\,
    p_{tr}(\mathcal{M}\mid [\mathcal{M}^o], do(V))\,
    d\mathcal{M}.
\end{equation}
\emph{Proof.}
Starting from the ACTIVA generative model in \Eqref{eq:activa_gen_process}, we marginalize out the latent variable:
\begin{equation}
    p_{\theta^*}(\boldsymbol{V}\mid \boldsymbol{D}^{\mathcal{M}^o}, do(V))
    =
    \int
    p_{\gamma^*}(\boldsymbol{V}\mid \boldsymbol{z})\,
    p_{\eta^*}(\boldsymbol{z}\mid \boldsymbol{D}^{\mathcal{M}^o}, do(V))\,
    d\boldsymbol{z}.
\end{equation}
Using Proposition~1, this becomes
\begin{align}
    p_{\theta^*}(\boldsymbol{V}\mid \boldsymbol{D}^{\mathcal{M}^o}, do(V))
    &=
    \int
    p_{\gamma^*}(\boldsymbol{V}\mid \boldsymbol{z})\,
    \bar{q}_{\phi^*}(\boldsymbol{z}\mid [\mathcal{M}^o], do(V))\,
    d\boldsymbol{z} \\
    &=
    \int
    \left[
        \int
        p_{\gamma^*}(\boldsymbol{V}\mid \boldsymbol{z})\,
        q_{\phi^*}(\boldsymbol{z}\mid \mathcal{\mathbf{D}}^{tr})\,
        d\boldsymbol{z}
    \right]
    p_{tr}(\mathcal{\mathbf{D}}^{tr}\mid [\mathcal{M}^o], do(V))\,
    d\mathcal{\mathbf{D}}^{tr}.
\end{align}
By Assumption~2, the inner integral equals the true post-interventional distribution of the model that generated the task $\mathcal{\mathbf{D}}^{tr}$:
\begin{equation}
    \int
    p_{\gamma^*}(\boldsymbol{V}\mid \boldsymbol{z})\,
    q_{\phi^*}(\boldsymbol{z}\mid \mathcal{\mathbf{D}}^{tr})\,
    d\boldsymbol{z}
    =
    p_{\mathcal{M}}(\boldsymbol{V}\mid do(V)).
\end{equation}
Substituting this into the previous expression gives an integral over tasks whose integrand depends on the task only through the post-interventional distribution of the model that generated it. Since the simulator draws a model and then samples the task from it, the conditional task distribution factorizes as
% New display (added together with the surrounding \replaced/\added text): the factorization of the task distribution over models.
\begin{equation}\label{eq:prop2_task_factorization}
    p_{tr}(\mathcal{\mathbf{D}}^{tr}\mid [\mathcal{M}^o], do(V))
    =
    \int
    p_{tr}(\mathcal{\mathbf{D}}^{tr}\mid \mathcal{M}, do(V))\,
    p_{tr}(\mathcal{M}\mid [\mathcal{M}^o], do(V))\,
    d\mathcal{M}.
\end{equation}
Substituting this factorization and exchanging the order of integration, the integrand depends on the task only through its generating model $\mathcal{M}$, which is fixed in the inner integral, so it can be taken outside:
% New display (added together with the surrounding \added text): the reordered integrals.
\begin{align}
    &p_{\theta^*}(\boldsymbol{V}\mid \boldsymbol{D}^{\mathcal{M}^o}, do(V)) \notag \\
    &=
    \int
    p_{\mathcal{M}}(\boldsymbol{V}\mid do(V))
    \left[
        \int
        p_{tr}(\mathcal{\mathbf{D}}^{tr}\mid \mathcal{M}, do(V))\,
        d\mathcal{\mathbf{D}}^{tr}
    \right]
    p_{tr}(\mathcal{M}\mid [\mathcal{M}^o], do(V))\,
    d\mathcal{M}.
\end{align}
The inner integral of the task density over all tasks generated by $\mathcal{M}$ equals one, which yields
\begin{equation}
    p_{\theta^*}(\boldsymbol{V}\mid \boldsymbol{D}^{\mathcal{M}^o}, do(V))
    =
    \int
    p_{\mathcal{M}}(\boldsymbol{V}\mid do(V))\,
    p_{tr}(\mathcal{M}\mid [\mathcal{M}^o], do(V))\,
    d\mathcal{M},
\end{equation}
which proves the claim. $\Box$

\section{Datasets}\label{apx:datasets}
We evaluate the performance of the proposed method across three different types of datasets: two purely synthetic (\emph{Gaussian} and \emph{Beta}) and one semi-synthetic (\emph{SERGIO}). Below, we provide an overview of each dataset category along with the relevant parameters. For every causal model in these categories, we draw $N$ samples from both the observational and the interventional distributions to create paired datasets. Details about the exact number of models, how we split into training/test sets, and further parameter choices are given in the subsections below.

\subsection{Synthetic Data (Gaussian and Beta Noise)}\label{apx:linear_datasets}
We generate data from linear additive causal models of the form:
$$
V_j = \sum_{i \in \mathrm{Pa}(j)} \beta_{ij}\, V_i + \varepsilon_j,
$$
where \(\varepsilon_j\) is drawn from either: A Gaussian distribution, \(\mathcal{N}(0,\sigma^2)\), or  A Beta distribution, \(\mathrm{Beta}(\alpha,\beta)\).
For each variable \(V_i\), we generate interventional data by applying a single-variable intervention $do(V_i = 5)$ plus some noise $\mathcal{N}(0, 0.1)$ to increase numerical stability. In total, this yields one observational dataset and $d$ interventional datasets for each causal model. All synthetic data is generated with the \emph{Causal Playground} library~\citep{Sauter2024CausalPlayground:Research}. We split each dataset into a train (80\%), test (10\%) and validation (10\%) set.

The causal graphs are generated according to the ER procedure \citep{Erdos1959OnGraphs} where first we generate an ER graph with edge-probability $0.3$ and then randomly remove edges until the graph becomes acyclic. When the noise terms follow a Gaussian distribution, we sample \(\beta_{ij}\) uniformly from \([-2, -0.5]\cup[0.5, 2]\) and let \(\varepsilon_j\sim\mathcal{N}(0,0.5)\). For the Beta case, we again draw \(\beta_{ij}\) in \([-2, -0.5]\cup[0.5, 2]\), but the noise terms \(\varepsilon_j\) follow \(\mathrm{Beta}(\alpha,\beta)\) with \(\alpha,\beta\sim\mathrm{Uniform}(0.5,2)\). We examine both a 2-variable and an 8-variable scenario: in the 2-variable case, we randomly generate 3000 linear models and sample 100 points each for the observational and interventional distributions, whereas in the 8-variable case, we generate 20000 linear models and, for each model, sample 80 data points for the observational and for each interventional distribution.

\subsection{Semi-Synthetic Data (SERGIO)}

\label{apx:SERGIO_explaination}
To evaluate on data with biological realism, we employ the SERGIO simulator~\citep{Dibaeinia2020SERGIO:Networks}, which generates single-cell gene-expression data. Notably, we use the version of SERGIO provided in \citep{Lorch2022AmortizedLearning}, which includes functionality for performing interventions. We treat single-gene knockouts as interventions, thus applying $do(V_i=0)$ to each gene \(V_i\). This yields one observational dataset and 8 single-gene interventional datasets per simulated gene-regulatory network.

\textbf{Simulator Settings and Network Structures.}
Following the procedure in~\citep{Marbach2009GeneratingMethods,Lorch2021DiBS:Learning}, we randomly sample subgraphs of known gene-regulatory networks from \emph{E.~coli} or \emph{S.~cerevisiae}, ensuring that each subgraph has 8 genes. For each subgraph, we draw model parameters (e.g., activation constants, decay rates, and noise magnitudes) from predefined ranges (see Table \ref{tab:SERGIO_parameters}). We then generate observational data as well as data from each gene-knockout intervention.

\textbf{Training, Testing, and Out-of-Distribution (OOD) Splits.}
We run 15{,}000 SERGIO simulations for our in-distribution dataset and sample 30 cells (data points) for each observational and interventional condition. We then split these 15{,}000 simulations into training (80\%), validation (10\%) and test (10\%). We also construct an OOD set of 1800 simulations, where some SERGIO parameters (e.g., noise or decay rates) are sampled from partially disjoint ranges, creating a controlled distribution shift. For evaluation, we again sample 30 cells per observational/interventional condition in these OOD simulations.

\begin{table}[]
    \centering
    
    \caption{SERGIO data generation parameters.}
    \begin{tabular}{c|cc}
    parameter& in-distribution&out-of-distribution\\\hline
        genes & 8 & 11\\
        b & Uniform(0, 1)&Uniform(0.5, 2.0) \\
        k\_param & Uniform(1, 5)&Uniform(3, 7)\\
        k\_sign&Beta(1, 1)&Beta(0.5, 0.5)\\
        hill&$\in$\{ 1.9, 2.0, 2.1\}&$\in$\{ 1.5, 2.5\}\\
        decays&$\in$\{ 0.7, 0.8, 0.9\}&$\in$\{ 0.5, 1.5\}\\
        noise\_params&$\in$\{0.9, 1.0, 1.1\}&$\in$\{0.5, 1.5\}\\
        \hline
    \end{tabular}
    \label{tab:SERGIO_parameters}
\end{table}

\section{MACE-TNP Extension Implementation Details}                
  \label{app:mace-tnp-implementation}                    
  This section describes our extension of MACE-TNP (Model-Averaged Causal Estimation Transformer Neural Process)~\citep{Dhir2025EstimatingMeta-Learning} in detail. MACE-TNP is a meta-learned neural process for causal   
  inference that predicts \emph{marginal} interventional distributions $p(x_i \mid \operatorname{do}(X_j = v), \mathcal{D}_\text{obs})$ for a designated       
  outcome variable~$X_i$, given an intervention on variable~$X_j$.                                                                                             
  Its architecture consists of: (1)~a Transformer-based encoder that alternates attention across the sample axis and the variable (node) axis to produce
  permutation-equivariant per-node representations, and (2)~a per-variable MoG predictor that extracts the representation at the outcome node index~$i$ and    
  decodes it into a univariate mixture of Gaussians with $K$ components.
  The encoder uses three distinct embedding roles---\emph{intervention node}, \emph{outcome node}, and \emph{marginalized node}---each implemented as a        
  separate MLP that maps scalar observations to $d_\text{model}$-dimensional vectors, with separate copies for observational and interventional samples (six   
  MLPs total).
  Sample-wise attention uses multi-head self-attention (MHSA) on observational tokens followed by multi-head cross-attention (MHCA) from interventional tokens 
  to observational tokens; node-wise attention uses MHSA across all nodes.                                                                                     
  The predictive distribution factorizes over interventional samples: $p_\theta(\mathbf{x}_i \mid \operatorname{do}(\mathbf{x}_j), \mathcal{D}_\text{obs}) =
  \prod_{n=1}^{N_\text{int}} p_\theta(x_i^n \mid \operatorname{do}(x_j^n), \mathcal{D}_\text{obs})$.                                                           
  %Since the original model requires specifying both an intervention index~$j$ and an outcome index~$i$ at each forward pass, evaluating the full joint  interventional distribution requires $D - 1$ separate forward passes (one per non-intervened outcome variable).                                              
                  
  Our implementation modifies MACE-TNP to estimate the \emph{joint} interventional distribution %$p(\mathbf{x} \mid \operatorname{do}(X_j = v))$ 
  over all %$D$   
  variables simultaneously, rather than per-variable marginals.
  This requires three architectural changes:                                                                                                                   
                                                                                                                                                               
  \subsection{Two-role encoder.}                                                                                                                                
  Since we predict the joint distribution over all variables, there is no designated outcome variable and hence no need for the outcome embedding role.        
  We replace the original three-role encoder with a two-role encoder that uses only \emph{marginalized} (variable) and \emph{intervention} embeddings.         
  For each node $d \in \{1, \ldots, D\}$, the embedding is:                                                                                                    
  \begin{equation}                                                                                                                                             
      \mathbf{h}_d = \begin{cases}                                                                                                                             
          f_{\text{int}}(x_d) & \text{if } d = j \text{ (intervened variable)}, \\                                                                             
          f_{\text{var}}(x_d) & \text{otherwise},                                                                                                              
      \end{cases}                                                                                                                                              
  \end{equation}                                                                                                                                               
  where $f_{\text{var}}$ and $f_{\text{int}}$ are learned MLPs (with separate copies for observational and interventional inputs, giving four MLPs total).     
  The Transformer encoder backbone is left unchanged.                                                                                                                               
                                                                                                                                                               
  \subsection{Joint multivariate GMM predictor.}                                                                                                                
  We replace the per-variable univariate GMM predictor with a multivariate predictor that outputs a $K$-component Gaussian mixture model via Cholesky factors. %over $\mathbb{R}^D$:
  % \begin{equation}                                                                                                                                             
  %     p(\mathbf{x} \mid \operatorname{do}(X_j = v)) = \sum_{k=1}^{K} \pi_k \, \mathcal{N}\!\left(\mathbf{x};\, \boldsymbol{\mu}_k,\, \mathbf{L}_k
  % \mathbf{L}_k^\top\right),                                                                                                                                    
  % \end{equation}  
  % where $\boldsymbol{\mu}_k \in \mathbb{R}^D$, $\mathbf{L}_k \in \mathbb{R}^{D \times D}$ is a lower-triangular matrix (the Cholesky factor of the covariance),
  %  and $\pi_k$ are mixture weights with $\sum_k \pi_k = 1$.                                                                                                    
  The predictor takes per-node encodings of shape, obtained by mean-pooling the encoder output over the target-sample axis.
  Component means are computed \emph{per-node}: each node's $d_\text{model}$-dimensional encoding is independently mapped to $K$ scalar means via a shared     
  two-layer MLP, preserving the per-variable causal information encoded by the Transformer.                                                                    
  For the Cholesky factors and mixture weights, the node encodings are mean-pooled across the variable axis and decoded through a shared backbone MLP followed 
  by separate linear heads.                                                                                                                                    
  The diagonal entries of each $\mathbf{L}_k$ are passed through a softplus activation with a floor of $0.1$ to ensure positive-definiteness, and mixture
  weights are obtained via softmax.                                                                                                                            
                  
  \subsection{Training and inference.}                                                                                                                          
  The model is trained by minimizing the negative log-likelihood under the joint GMM like in the original paper. %:
  % \begin{equation}                                                                                                                                             
  %     \mathcal{L} = -\frac{1}{BN} \sum_{b=1}^{B} \sum_{n=1}^{N} \log p\!\left(\mathbf{x}_{b,n} \mid \operatorname{do}(X_{j_b} = v_b)\right),
  % \end{equation}                                                                                                                                               
  % where $\mathbf{x}_{b,n}$ are the $N$ interventional target samples.
  During training, a single randomly sampled target token is used as the encoder query (matching the inference-time regime), while the loss is evaluated       
  against all target samples in the batch.                                                                                                                     
  At test time, a dummy target token is constructed with only the intervention variable set to its (z-score normalized) intervention value and the remaining   
  entries zeroed out; the encoder processes this alongside the observational context, and the predicted GMM is sampled to produce draws from the estimated   joint interventional distribution.
                                                                                                                                                               
  Both observational and interventional data are z-score normalized using the observational data statistics (per-sample mean and standard deviation), following
   the normalization convention of the original MACE-TNP.
  Predictions are denormalized before metric evaluation.                                                                                                       
  Training uses AdamW with cosine annealing, and gradient clipping (max norm $1.0$).                                  
  
  \subsection{Architectural symmetry properties.}                                                                                                               
  The original MACE-TNP is permutation-equivariant with respect to both observational samples and variables (nodes).
  The latter property also allows the model to generalize to a different number of variables at test time than seen during training, since the per-variable MoG
   predictor extracts a single node's representation independently of~$D$.                                                                                     
  Our joint formulation preserves the sample-axis symmetries: permutation invariance with respect to observational samples and permutation equivariance with   
  respect to interventional samples, since the underlying module is identical to the original.                            
                  
  However, our predictor head does preserve permutation equivariance only for the means, not for the covariances.                                        
  The component means~$\boldsymbol{\mu}_k$ are computed per-node, so they transform equivariantly
   under variable permutations.
  The Cholesky factors~$\mathbf{L}_k$, however, are decoded from a mean-pooled representation over the variable axis, which is permutation-\emph{invariant}.   
  Consequently, $\mathbf{L}_k$ is the same regardless of variable ordering.                                                                                                                                              
  This means the model implicitly assumes a fixed variable ordering for the covariance structure and must learn the correct assignment of covariance entries to
   variable pairs from training data.                                                                                                                          
  Furthermore, because the Cholesky head outputs a fixed number of parameters, the number of variables is determined at construction time and
  cannot vary at inference, unlike the original per-variable formulation.                                                                                      
                  
  These are deliberate trade-offs: they allow us to use efficient mean-pooling while predicting a full joint distribution. In our experimental setting, these constraints mainly affect the inductive bias and sample efficiency of the baseline rather than the validity of the maximum-likelihood training objective. We therefore view this extension as a practical joint-distribution adaptation of MACE-TNP, not as a claim that the original permutation-equivariance properties are fully preserved.

\section{Distribution Plots}\label{apx:distribution_plots}
\begin{figure}[htbp]
    \centering
    \includegraphics[width=0.85\textwidth]{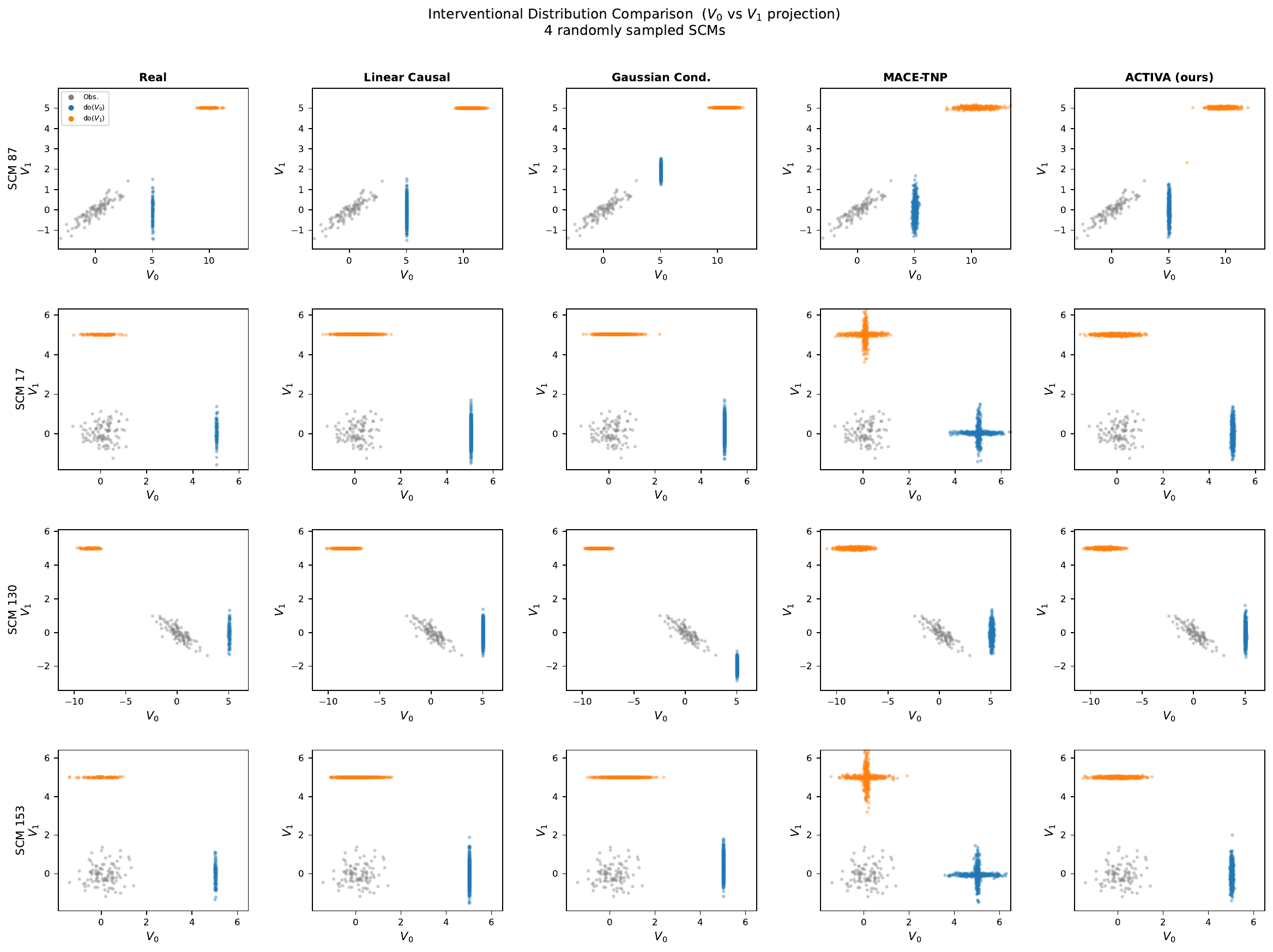}
    \caption{\emph{Gauss 2} test data.}
    \label{fig:dist_a}
\end{figure}

\begin{figure}[htbp]
    \centering
    \includegraphics[width=0.85\textwidth]{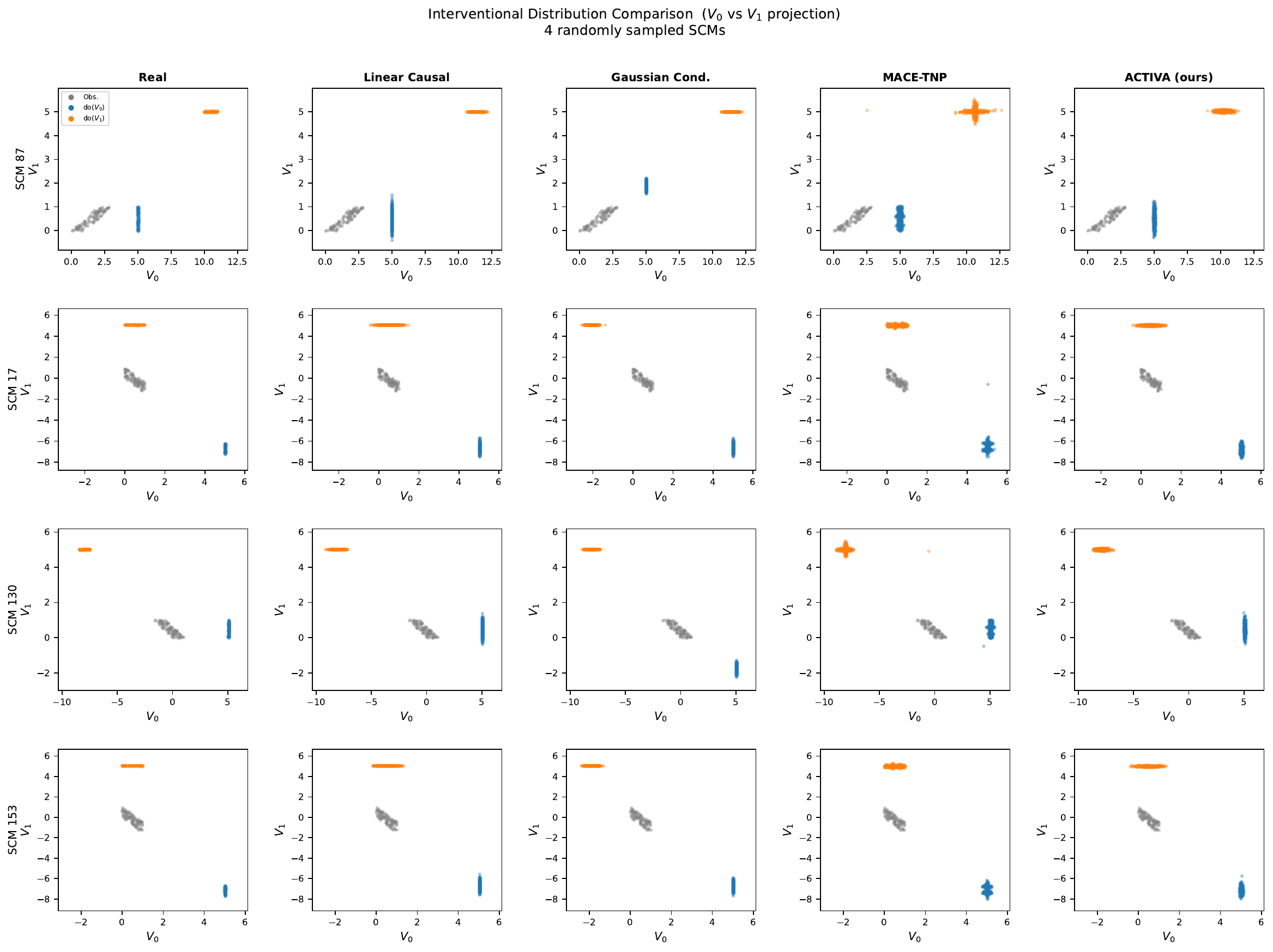}
    \caption{\emph{Beta 2} test data.}
    \label{fig:dist_b}
\end{figure}

\begin{figure}[htbp]
    \centering
    \includegraphics[width=0.85\textwidth]{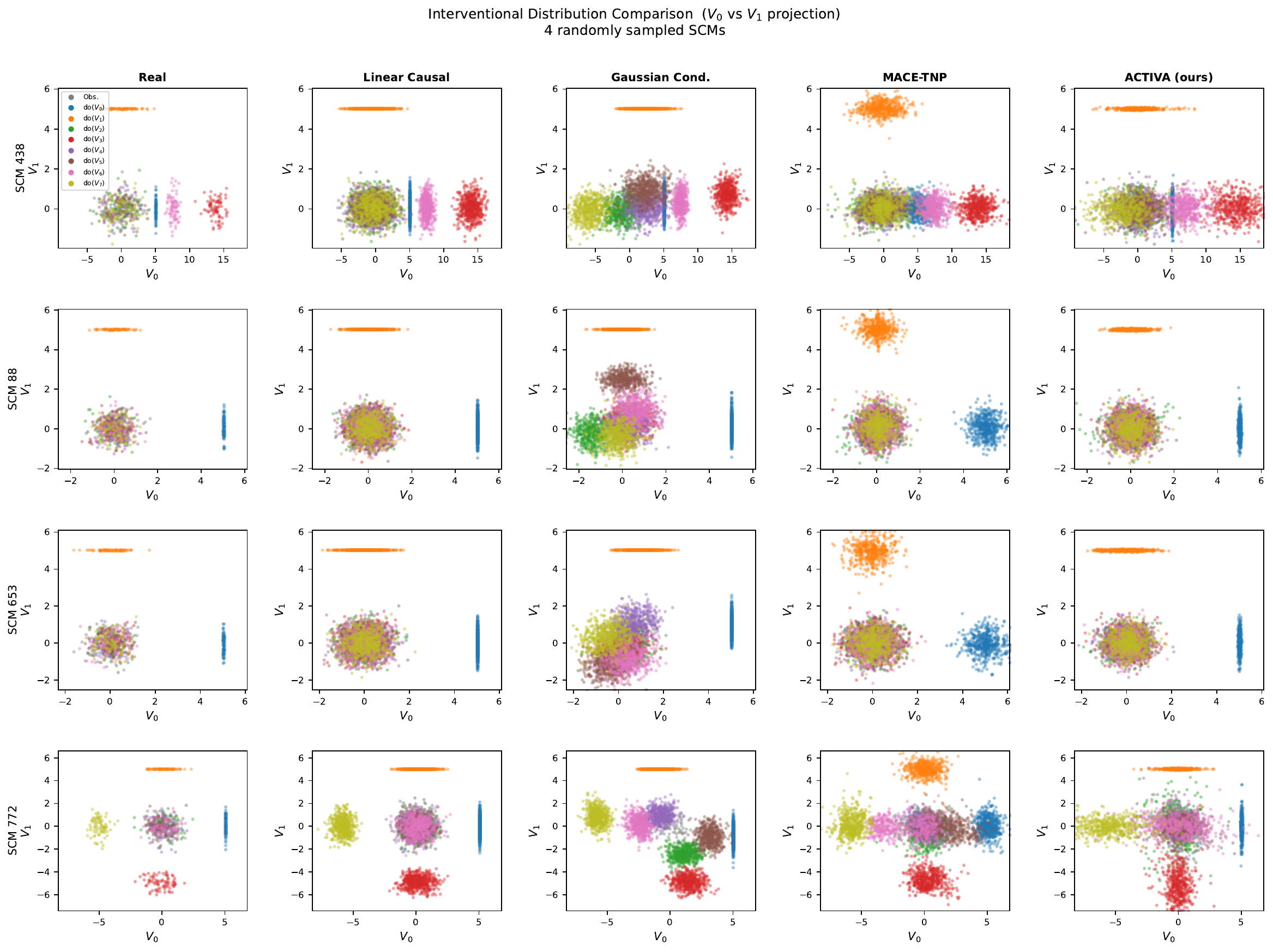}
    \caption{\emph{Gauss 8} test data.}
    \label{fig:dist_c_normal_peach_sun}
\end{figure}

\begin{figure}[htbp]
    \centering
    \includegraphics[width=0.85\textwidth]{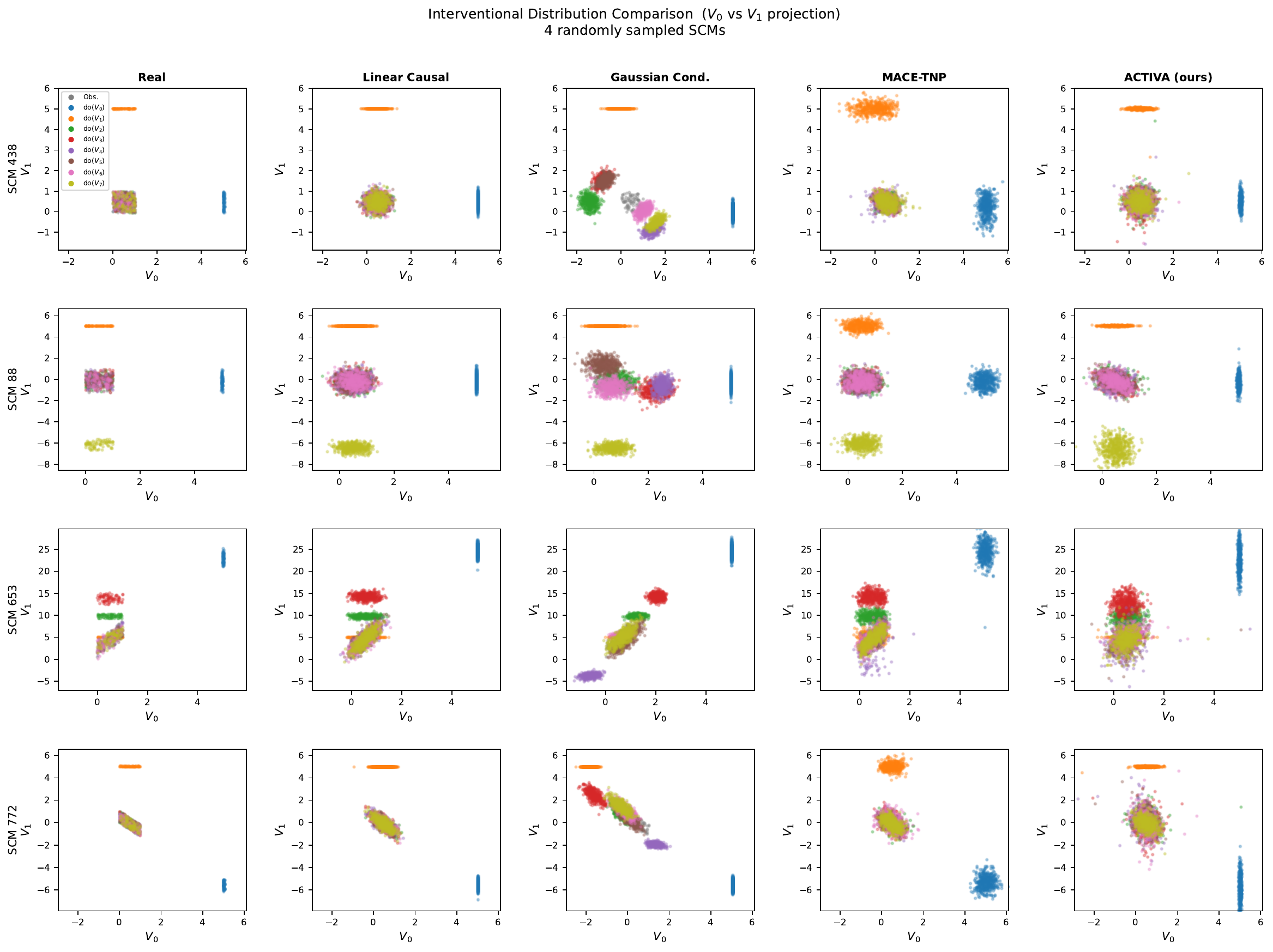}
    \caption{\emph{Beta 8} test data.}
    \label{fig:dist_c_beta_youthful_resonance}
\end{figure}

In the Figures~\ref{fig:dist_a}-~\ref{fig:dist_c_beta_youthful_resonance}, each column represents samples from the observational and interventional distributions of one model, while each row corresponds to a different randomly drawn SCM from the corresponding test set. The first column shows samples from the ground-truth SCM, columns two to four show the baselines, and the last column shows our model.

Within each panel, the gray points correspond to the true observational data and are repeated across all plots to serve as a common reference. Colored points represent samples from interventional distributions, where the color encodes which variable $V_i$ has been intervened on. All samples are projected onto the $(V_0, V_1)$ plane, so differences between models are visible as shifts or distortions of the colored point clouds relative to the gray observational background and to the ground-truth column.

Qualitatively, a model is accurate if, for each intervention (color), the resulting cloud of points in the corresponding column closely matches the shape, location, and spread of the ground-truth interventional samples in the first column. Systematic deviations in any of these aspects indicate characteristic failure modes: for instance, collapsing the spread suggests underestimation of uncertainty, while consistent shifts or rotations of the point clouds signal misspecified causal effects. By comparing these patterns row-wise across SCMs and column-wise across methods, we can visually assess how reliably each approach captures the interventional behavior of the underlying causal model from observational data.
%----------------- dist_a -----------------

%----------------- dist_b -----------------

%----------------- dist_c_1 -----------------

%----------------- dist_c_2 -----------------

%----------------- dist_c_3 -----------------
% \begin{figure}[htbp]
%     \centering
%     \includegraphics[width=0.85\textwidth]{figures/dist_comparison_8var_sergio_nodrop_comfy_sea_best.pdf}
%     \caption{\emph{SERGIO} test data.}
%     \label{fig:dist_c_sergio_nodrop_comfy_sea}
% \end{figure}

% If you want to avoid naming each file manually and are using latexmk or a build system,
% you still need to list the files explicitly in LaTeX; wildcard expansion like dist_*.pdf
% is not supported directly in \includegraphics.

\section{LLM Usage} %TODO discuss
Throughout this work, we made use of various large language models (LLMs) as general-purpose assistive tools across different stages of the research process. Specifically, LLMs were employed to support literature exploration, clarify technical formulations, check the consistency of proofs, assist with coding and debugging, suggest concise rephrasings for improved readability, and aid with formatting tasks (e.g., LaTeX adjustments). All scientific contributions, conceptual developments, and final claims remain the responsibility of the authors.

\end{document}